\documentclass[a4paper,fleqn]{cas-sc}

\usepackage[numbers]{natbib}
\usepackage{amssymb}
\usepackage{multirow}
\usepackage{algorithmicx}
\usepackage[ruled]{algorithm2e}
\usepackage{amsfonts}
\usepackage{amsmath}
\usepackage{amsthm}
\usepackage{balance}
\usepackage{bm}
\usepackage{bbm}
\usepackage{booktabs} % for professional tables
\usepackage{color}
\usepackage{enumitem}
\usepackage{float}
\usepackage{graphicx}
\usepackage{hyperref}
\usepackage{makecell}% Yue add this package 
\usepackage{microtype}
\usepackage{multirow}
\usepackage{natbib}
\usepackage{xcolor}
\usepackage{siunitx}
\usepackage{stfloats}
\usepackage{subfigure}
\usepackage[flushleft]{threeparttable}
\usepackage{url}
\usepackage{array}
\usepackage{booktabs}
\usepackage{enumitem}
\usepackage{caption}
\usepackage{subcaption}
\makeatletter
\renewcommand{\fnum@figure}{Fig. \thefigure.\@gobble}
\makeatother

\definecolor{softgreen}{RGB}{34, 139, 34} 

\def\tsc#1{\csdef{#1}{\textsc{\lowercase{#1}}\xspace}}
\tsc{WGM}
\tsc{QE}
\begin{document}

\let\printorcid\relax
\let\WriteBookmarks\relax
\def\floatpagepagefraction{1}
\def\textpagefraction{.001}

% \shorttitle{<short title of the paper for running head>}    
\shorttitle{CureGraph: Contrastive Multi-Modal Graph Representation Learning for Urban Living Circle Health Profiling and Prediction}   

% \shortauthors{<short author list for running head>} 
%\shortauthors{Jinlin Li et al.}
\shortauthors{J.Li and X.Zhou}

\title[mode = title]{CureGraph: Contrastive Multi-Modal Graph Representation Learning for Urban Living Circle Health Profiling and Prediction}

\author[1]{Jinlin Li}[style=chinese]

\author[1]{Xiao Zhou}[style=chinese]
\cormark[1] 
\ead{xiaozhou@ruc.edu.cn}

\address[1]{Gaoling School of Artificial Intelligence, Renmin University of China, Beijing, China}
\cortext[1]{Corresponding author at: Gaoling School of Artificial Intelligence, Renmin University of China, Beijing, China.}

\begin{abstract}
The early detection and prediction of health status decline among the elderly at the neighborhood level are of great significance for urban planning and public health policymaking. While existing studies affirm the connection between living environments and health outcomes, most rely on single data modalities or simplistic feature concatenation of multi-modal information, limiting their ability to comprehensively profile the health-oriented urban environments.
To fill this gap, we propose \textbf{CureGraph}, a \textbf{c}ontrastive m\textbf{u}lti-modal \textbf{r}epresentation learning framework for urban h\textbf{e}alth prediction that employs \textbf{graph}-based techniques to infer the prevalence of common chronic diseases among the elderly within the urban living circles of each neighborhood.
CureGraph leverages rich multi-modal information, including photos and textual reviews of residential areas and their surrounding points of interest, to generate urban neighborhood embeddings. By integrating pre-trained visual and textual encoders with graph modeling techniques, CureGraph captures cross-modal spatial dependencies, offering a comprehensive understanding of urban environments tailored to elderly health considerations. Extensive experiments on real-world datasets demonstrate that CureGraph improves the best baseline by $28\%$ on average in terms of $R^2$ across elderly disease risk prediction tasks.
Moreover, the model enables the identification of stage-wise chronic disease progression and supports comparative public health analysis across neighborhoods, offering actionable insights for sustainable urban development and enhanced quality of life. The code is publicly available at \href{https://github.com/jinlin2021/CureGraph}{https://github.com/jinlin2021/CureGraph}.
\end{abstract}
%\includegraphics{}red

%\item highlight-1
%\item highlight-2
%\item highlight-3
% Keywords
\maketitle

\section{Introduction}
\label{sec:sample1}
~\

Over the past two decades, rapid urbanization and the global trend of an ageing population have emerged as  inevitable and intersecting demographic forces shaping the current century. According to statistics~\cite{rudnicka2020world}, no country had more than 18 percent of its population over the age of 65 in 2000. However, due to the accelerating ageing process, this proportion is projected to reach 38 percent by 2050. This demographic shift would inevitably pose a series of significant challenges to cities worldwide, underscoring the need for more age-friendly urban environments. Among the various factors influencing the quality of life for the elderly, chronic conditions that increase with age are critical and must not be overlooked. In 2020, the UN General Assembly declared 2021–2030 as the Decade of Healthy Aging~\cite{world2020decade}, highlighting the urgent need for policymakers worldwide to prioritize initiatives aimed at enhancing the well-being of older individuals. Therefore, devising an effective health-oriented profiling and prediction framework for aging-friendly urban environments—grounded in a deep understanding of the linkages between the multi-modal semantics of neighborhood living circles and chronic diseases associated with aging—is essential in this era of longevity and forms the core research focus of this paper.

In current literature, a substantial body of research has revealed that the characteristics of human settlements are intricately associated with the chronic health conditions of urban senior citizens. For instance,~\citeauthor{cassarino2015environment}~\cite{cassarino2015environment} demonstrated that a community's social and built environment can influence elderly cognition, with complex neighborhood settings potentially delaying cognitive decline. A study on Hong Kong communities~\cite{guo2021objective} found that the relationship between street connectivity and mental health follows an inverted U-shaped pattern, while park green spaces and mixed land use positively impact mental health and subjective well-being.~\citeauthor{wang2019relationship}~\cite{wang2019relationship} utilized the street view images to assess how neighborhood walkability factors-such as sky visibility, greenery, and building proportion-affect elderly mental health, particularly regarding depression and anxiety.  Similarly,~\citeauthor{song2022using}~\cite{song2022using} found that green and blue spaces, population density, and public transport accessibility present statistically significant relationships with blood pressure. ~\citeauthor{maharana2018use}~\cite{maharana2018use} applied a convolutional neural network (CNN) to extract built environment features from high-resolution satellite images, exploring links between physical attributes (e.g., parks, highways, green streets, pedestrian crossings, and housing types) and the health conditions of community residents.

In addition to the physical environment of neighborhoods, another line of research has focused on the connections between lifestyle choices and residents' health status. For instance,~\citeauthor{li2013prevalence}~\cite{li2013prevalence} examined the prevalence and potential risk factors of mild cognitive impairment (MCI), highlighting that a healthy diet, participation in physical, intellectual, and social activities, and effective management of vascular-related diseases are crucial for maintaining cognitive and brain health.~\citeauthor{galiatsatos2018neighbourhood} ~\cite{galiatsatos2018neighbourhood} investigated the relationships between socio-economic determinants, tobacco store density, and health outcomes at the neighborhood level. Similarly, ~\citeauthor{hasthanasombat2019understanding}~\cite{hasthanasombat2019understanding} analyzed the causal impact of sports venue availability on the prevalence of antidepressant prescriptions in London neighborhoods.~\citeauthor{ijcai2018p497}~\cite{ijcai2018p497} utilized check-in data from various points of interest (POIs) to capture local lifestyle patterns within neighborhoods and predict the progression rates of chronic diseases.

Although there is a multitude of studies advocate for actions on the built environment and lifestyle choices to improve health outcomes for the elderly, the evidence base for these findings often relies on urban data from a single modality (e.g., social demography, street view images, or textual reviews on social networks) to explore relationships between specific neighborhood characteristics and particular diseases. A health-focused neighborhood representation learning framework that comprehensively considers the scope of the elderly's daily activities and leverages rich multi-modal urban data to predict age-related health declines is still lacking.
To address this gap, this study proposes a contrastive multi-modal graph representation learning approach for profiling the health of the elderly in urban communities. The novelty of this research lies in three key aspects. First, from the perspective of \textit{spatial scope}, it offers a contextualized view on healthy aging-focused built environment studies by jointly analyzing both intra-community and inter-community conditions. Second, in terms of \textit{methodology}, it utilizes multi-modal urban data and introduces fusion techniques for hierarchical neighborhood representation learning. Finally, in \textit{practical terms}, it provides actionable recommendations for public health practitioners and urban planners to enhance spatial planning for healthy aging.

\begin{figure}[t!]
	\centering
	\includegraphics[width=0.6\textwidth]{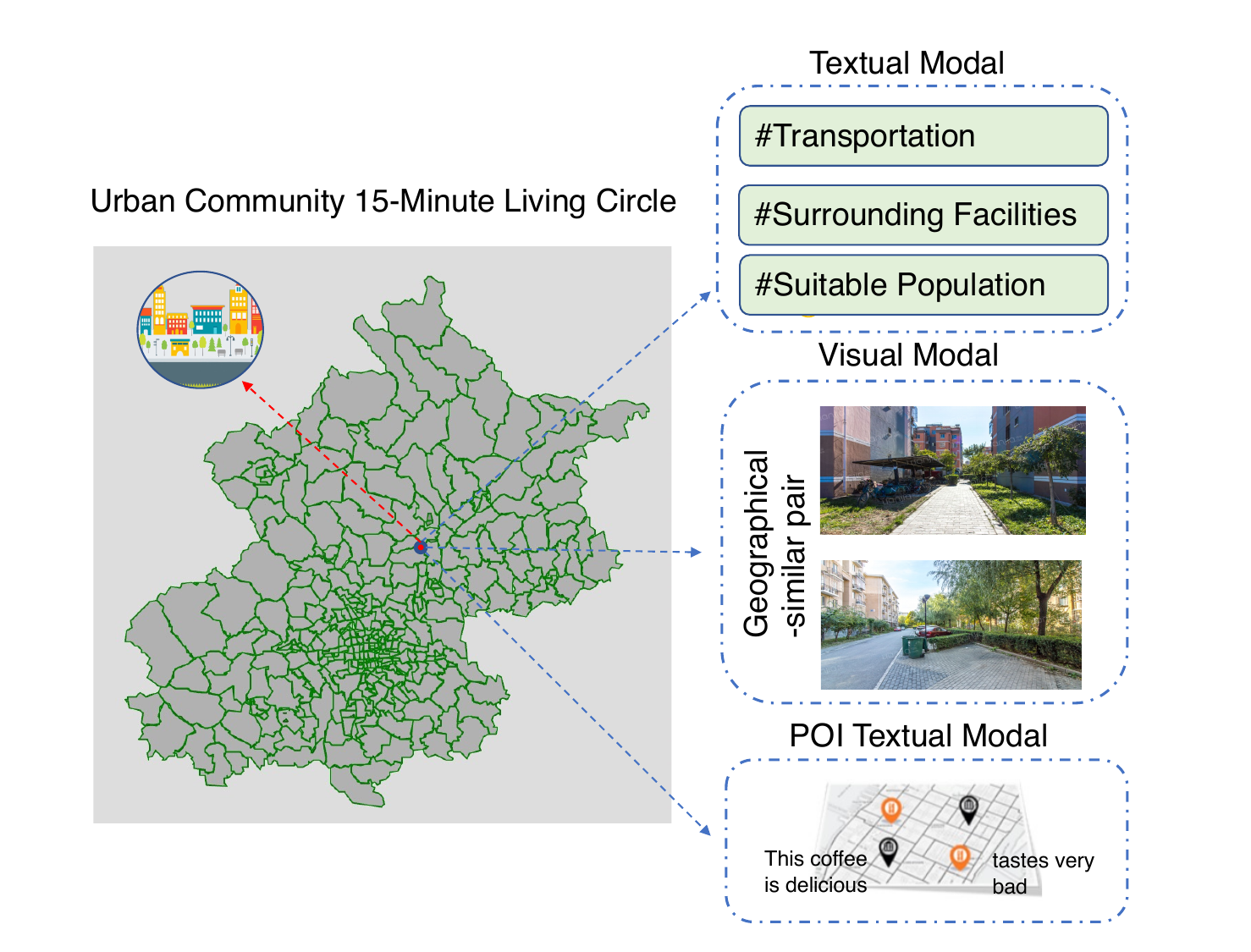}
	\caption{Multi-modal data of urban community 15-minute living circles.}
	%\vspace{-2ex}
	\label{fig:multimodal_bj}
	%\vspace{-2ex}
\end{figure}

More specifically, since neighborhoods serve as the primary venue for the elderly to carry out their daily activities and are a crucial determinant of their quality of life—particularly due to declining physical functions, social adaptability, and increased environmental vulnerability~\cite{wiles2009older,wiles2012meaning, aneshensel2016aging}—we introduce the concept of the \textit{urban community living circle}~\cite{moreno2021introducing,ma202315} as the core unit of study. In this framework, the 15-minute living circle surrounding each residential area is considered the main activity space for elderly residents. This circle not only encompasses the physical elements within the residential district itself but also includes surrounding facilities that can be accessed within a 15-minute walk or bike ride. This approach enables us to account for both the physical environment of the neighborhood and the lifestyles shaped by nearby amenities. To fully capture the scope of the community living circles, we collect multi-modal datasets, including textual descriptions, reviews, and images of the target residential area and the POIs within each living circle. For better illustration, we present the distribution of the multi-modal data within the 15-minute community living circle in Beijing, as shown in Fig.~\ref{fig:multimodal_bj}.

We leverage pre-trained models to design contrastive learning tasks, utilizing a combination of geographic distance and data augmentation techniques. First, we use CNN-based methods and contrastive learning tasks based on geographic distance to extract visual features. Next, we introduce a cross-modal contrastive learning task to learn descriptive features from long-text descriptions within the living circle. Simultaneously, we apply supervised contrastive learning to extract review features from POIs, incorporating sentiment analysis as an additional layer of comparison. The features learned across modalities are then mapped to a common space via the modal encoder. To account for the spatial dependencies between urban areas, we propose an empirical measure—namely, the spatial autocorrelation matrix—to capture the geographical relationships and similarities in POI distribution between community living circles. By adopting this strategy, our framework can simultaneously integrate multi-modal data from the living circles while preserving local spatial constraints. Finally, through Graph Convolutional Network (GCN) encoding, with residual links and node splicing across modalities, we generate a representation for each living circle to predict the prevalence of chronic diseases among the elderly. The resulting graph embeddings also enable comparative health analyses across urban regions at multiple levels.

The contributions of this work are summarized as follows:
\begin{itemize}
    \item To the best of our knowledge, this is the first study to leverage multi-modal information with local constraints from urban living circles to address health issues among the elderly.
    \item We propose a novel contrastive multi-modal graph-based model for urban area representation learning, capturing both inter-modality interactions across different living circles and intra-modality interactions within each circle.
    \item By introducing a spatial autocorrelation matrix for community living circles, our approach provides new insights into utilizing the spatial characteristics of living circle elements to predict aging-related chronic diseases.
    \item The proposed framework is highly flexible and versatile, capable of uncovering urban health disparities across multiple dimensions. It also facilitates the integration of additional neighborhood features (e.g., housing prices, satellite imagery) to enrich embedding learning.
    \item We validate the effectiveness of the proposed framework using real-world urban datasets. Ablation studies further confirm the benefits of incorporating multi-modal information for neighborhood representation learning, contributing to healthy, age-friendly community planning.
\end{itemize}

\section{Related work}
\subsection{Contrastive Learning in Urban Settings}
~\

Inspired by significant advancements in self-supervised learning within computer vision~\cite{chen2020simple, he2020momentum, wang2022self} and natural language processing~\cite{wu2020clear, gao2021simcse, yan2021consert}, researchers have begun to extend these methodologies to urban modal coding. Despite these innovations, much of the existing research in this domain remains grounded in Tobler's First Law of Geography~\cite{miller2004tobler} when designing contrastive learning similarity measures. Self-supervised learning, particularly contrastive learning, has been widely used to distill expressive modal elements in urban spaces. 
For example,~\citeauthor{jean2019tile2vec}~\cite{jean2019tile2vec} employed triplet loss to minimize the distance between representations of spatially proximate satellite images while maximizing the distance for spatially distant pairs. Similarly,~\citeauthor{wang2020urban2vec}~\cite{wang2020urban2vec} utilized a comparable loss function for street view imagery.~\citeauthor{li2022predicting}~\cite{li2022predicting} applied the InfoNCE loss~\cite{oord2018representation} from the SimCLR framework~\cite{chen2020simple} to extract visual representations from satellite and street view images.~\citeauthor{xi2022beyond}~\cite{xi2022beyond} expanded this approach by incorporating a POI-based similarity metric, ensuring that images associated with similar POI category distributions yielded closer visual representations.~\citeauthor{liu2023knowledge}~\cite{liu2023knowledge} further extended this line of research by integrating a knowledge graph (KG) to represent urban knowledge within satellite imagery, formulating an image-KG contrastive loss that builds upon the traditional InfoNCE loss~\cite{oord2018representation}. 

However, the aforementioned studies rely solely on contrastive learning frameworks based on single-modality data or perspectives, overlooking the complementary potential of combining urban images and textual information.

\subsection{Urban Region Representation Learning}
~\

Urban region representation learning has garnered significant attention in recent years, driven by the proliferation of urban big data, the rapid advancements in embedding techniques within deep learning, and the pressing need for applications in smart cities. Broadly, research in this area often leverages word embedding or graph embedding techniques, tailored to address the spatio-temporal characteristics unique to urban settings.

For example,~\citeauthor{wang2017region}~\cite{wang2017region} constructed a mobility graph using fine-grained taxi commuting data and learned region embeddings at different time intervals. Similarly,~\citeauthor{yao2018representing}~\cite{yao2018representing} used taxi trajectories to learn embeddings of urban functional zones, applying the word2vec~\cite{mikolov2013distributed} method by treating each urban region as a word and mobility events as context. More recently, the urban computing community has increasingly recognized the importance of exploiting multi-modal urban data for region representation learning, given the complexity of cities as systems. A notable example is Urban2Vec by~\citeauthor{wang2020urban2vec}~\cite{wang2020urban2vec}, a neighborhood embedding method based on word embedding techniques that integrates multi-modal data, including street view images and POI textual reviews. From a multi-view perspective,~\citeauthor{fu2019efficient}~\cite{fu2019efficient} incorporated spatial relationships between regions, constructing multi-view POI-POI networks to represent urban regions.~\citeauthor{zhang2021multi}~\cite{zhang2021multi} advanced this approach by designing a cross-view information-sharing strategy and a weighted multi-view fusion technique to combine human mobility and POI data. Similarly,~\citeauthor{huang2021learning}~\cite{huang2021learning} utilized multimodal data as node and edge features within a multi-graph, leveraging human mobility data to represent neighborhood relationships while employing contrastive sampling to learn neighborhood representations.~\citeauthor{li2023urban}~\cite{li2023urban} proposed RegionDCL, an unsupervised framework utilizing OpenStreetMap (OSM) building footprints and POI data to learn regional representations. This approach incorporates dual contrastive learning at both the group and region levels, preserving spatial proximity and architectural similarity among building clusters.~\citeauthor{yong2024musecl}~\cite{yong2024musecl} proposed the MuseCL framework, which leverages contrastive learning across visual and textual modalities, combined with a cross-modality attentional fusion module, to enable fine-grained urban region profiling and socioeconomic prediction.~\citeauthor{xu2024cgap}~\cite{xu2024cgap} introduced the CGAP algorithm, incorporating hierarchical graph pooling with local and global attention mechanisms to improve urban region representation by integrating multi-modal data such as POIs and inter-regional human mobility patterns.

Given that graphs are well-suited for organizing urban elements and their complex relationships, and graph neural network (GNN)-based approaches have recently emerged as powerful tools for graph embedding tasks~\cite{velickovic2017graph,gilmer2017neural}, this study adopts graph convolutional networks (GCNs) as the base model to encode multi-modal graphs encompassing diverse urban data modalities. 
Additionally, our framework integrates advanced techniques, including contrastive learning, image and text analysis, and spatial pattern mining. This combination of methods distinguishes our community living circle representation learning framework from existing approaches, offering a novel and comprehensive solution.

\subsection{Health in Urban Environments}
~\

As the introduction section provides a comprehensive overview of studies related to health in urban built environments, we will not repeat them here. A key limitation in the current literature is the frequent neglect of non-linear interactions among urban features, which can significantly influence community health outcomes. Many conventional studies that employed traditional spatial statistics methods~\cite{liu2017relationships, maharana2018use, wang2019relationship, tang2022effect} assumed linear relationships between features, overlooking the inherent complexity and subtle dynamics of urban environments. In reality, urban health outcomes are influenced not only by the geographic distribution of facilities and resources, as described by Tobler’s First Law of Geography, but also by the complex interactions among various neighborhood characteristics, including social~\cite{huang2023reconstructing,zhou2023phase}, economic, mobility~\cite{zhou2018discovering}, and environmental factors. This complexity is consistent with the Second Law of Geography, which highlights spatial heterogeneity~\cite{anselin1989special}.

To address these limitations, our proposed method captures the relationships between feature modalities within urban living circles while accounting for spatial heterogeneity across different circles. By leveraging advanced techniques such as contrastive learning, graph representation learning, and spatial analysis, our approach provides a more nuanced understanding of the intricate interplay between factors influencing urban health outcomes at multiple scales. Additionally, it offers valuable insights for urban planners and policymakers, guiding strategies to improve the health and well-being of elderly populations in urban communities.

\section{Problem statement}\label{sec:prob_state}

~\

In this section, we delineate the related notations and the problem we study.

~\\
\textit{\textbf{Definition 1} (Urban Living Circle)}:
The health condition of a metropolitan area can be reflected by a series of urban living circle scenes. Among these, the community living circle serves as a fundamental spatial unit capable of meeting the essential daily activity requirements of its residents. These activities typically have high frequency, short duration, and are confined to the residential area and its immediate vicinity. In this study, we define the urban living circle as the service radius of a residential area, specifically referring to a 15-minute walking distance (within a 1000-meter radius) from various essential service facilities. This concept is commonly referred to as the "15-minute living circle," which also serves as the basic spatial unit for our downstream community health predictions. Given a set of urban living circles, $\mathcal{C}=\left\{c_{1}, c_{2}, \ldots, c_{n}\right\}$, each urban living circle $c_i$ contains a paragraph of residential area text, $\mathcal{T}_i$, a set of residential area images, ${\mathcal{V}_i} = \left\{v_{i1}, v_{i2}, \ldots, v_{iM_{i}}\right\}$, and a set of health-related POIs, ${\mathcal{P}_i} = \left\{p_{i1}, p_{i2}, \ldots, p_{i{O_i}}\right\}$ located within $c_i$. Here, $n$ represents the number of residential areas, and $M_{i}$ and $O_{i}$ denote the number of images and POIs contained within each living circle $c_i$, respectively. We assume that the GPS coordinates of the images and the residential area are consistent.

\begin{table}[htbp]
\centering
\caption{Descriptions of frequently used notations.}
\label{table:notation data}
\begin{tabular}{l|l}
\toprule
\bf Notation    
& \bf Description  \\ \midrule
  $\boldsymbol{c_i^v}$    &    The corresponding visual feature of living circle $c_i$.
\\ \hline
  $\boldsymbol{c_i^t}$    &    The corresponding textual feature of living circle $c_i$.
\\ \hline
  $\boldsymbol{c_i^p}$    &  The corresponding POI textual feature of living circle $c_i$.
\\ \hline
 $\operatorname{sim}(\cdot)$     &    The cosine similarity.
\\ \hline
 ${M_i}$      &     Number of images contained within each living circle $c_i$.
\\ \hline
   $\boldsymbol{e_i}$   &    $e_{i}\in\mathbb{R}^{d}$ is the embedding of $c_i$.
\\ \hline
   $\tau$   &    $\tau\in{R} $ is the temperature hyperparameter in
   modal encoder.
\\ \hline
   $S_{i,j}$   &   The spatial autocorrelation coefficients between $c_i$ and $c_j$.
\\ \hline
   $C_i^{Top-K}$   &    The top-$k$ most correlated candidate regions for living circle $c_i$.
\\ 
\bottomrule
\end{tabular}
\end{table}

~\\
\textit{\textbf{Definition 2} (Urban Living Circle Texts)}:
The living circle text describes the supporting facilities, transportation options, surrounding amenities, and the target residents of the residential area. The representation of urban living circles texts, denoted as $\mathcal{T}$, is formalized as follows:

\begin{eqnarray}
\mathcal{T}=\left\{\vec{c}_{1}^{t}, \vec{c}_{2}^{t}, \ldots, \vec{c}_{i}^{t}\right\}, \vec{c} \in \mathbb{R}^{{F}_{t}}, \forall \vec{c^{t}}\in\mathcal{T},
\end{eqnarray}
where ${c}_{i}^{t}$ is the corresponding textual feature of $i$-th region and ${F}_{t}$ is the number of dimensions of the feature.

~\\
\textit{\textbf{Definition 3} (Urban Living Circle Images)}:
The visual modality of the living circle captures the physical environment within the residential area and serves as a visual complement to the textual modality. The number of community images in each living circle may vary. We denote the set of urban living circle images as $\mathcal{V}$, defined as follows:
\begin{eqnarray}
\mathcal{V}=\left\{\vec{c}_{1}^{v}, \vec{c}_{2}^{v}, \ldots, \vec{c}_{i}^{v}\right\}, \vec{c} \in \mathbb{R}^{{F}_{v}}, \forall \vec{c^{v}}\in\mathcal{V},
\end{eqnarray}
where ${c}_{i}^{v}$ is the corresponding visual feature of $i$-th region and ${F}_{v}$ is the number of dimensions of the feature. 

~\\
\textit{\textbf{Definition 4} (Urban Living Circle POI Texts):}
POIs serve as direct representations of urban functions. The distinctive features of check-ins and review information associated with different types of POIs can be leveraged to infer the healthiness of lifestyles. We map the relevant POIs to their corresponding living circle locations and denote the set of urban living circle POI texts as $\mathcal{P}$, defined as follows:
\begin{eqnarray}
\mathcal{P}=\left\{\vec{c}_{1}^{p}, \vec{c}_{2}^{p}, \ldots, \vec{c}_{i}^{p}\right\}, \vec{c} \in \mathbb{R}^{{F}_{p}}, \forall \vec{c^{p}} \in \mathcal{P},
\end{eqnarray}
where ${c}_{i}^{p}$ is the corresponding POI textual feature of the $i$-th region, and ${F}_{p}$ is the dimensionality of the feature. Each POI is represented by textual data, including texts extracted from its categories and check-in reviews.

~\\
\textit{\textbf{Definition 5} (Spatial Correlation)}: The spatial correlation of a living circle is represented by the spatial autocorrelation matrix. This matrix is a variant of the covariance matrix of regions, capturing the pairwise similarity between each pair of 15-minute living circles.
\begin{eqnarray}
\mathcal{H}=\{\vec{h}_i\mid i=1,2,\ldots,n\}, \vec{h}_i=(c_1,c_2,\ldots,c_{|\vec{h}_i|}),
\end{eqnarray}
where $\overrightarrow{h_{i}}$ is a spatial neighborhood vector for an urban living circle $c_i$.

~\

\noindent \textit{\textbf{Problem Statement}}. Given a set of urban living circle scenes, $\mathcal{C}$, along with the corresponding residential area texts $\mathcal{T}$, visual data $\mathcal{V}$, and POI textual data $\mathcal{P}$, the main objective of the proposed model is to learn a distributed, low-dimensional health embedding, $\vec{e}$, for each urban living circle. Using these learned health embeddings as input, the model aims to predict the prevalence of four common geriatric diseases among the elderly within each urban living circle, namely Mild Cognitive Impairment (MCI), hypertension, diabetes, and Major Depressive Disorder (MDD). Additionally, the model seeks to generate health profiles at various spatial levels.

For brevity, frequently used notations are listed in Table~\ref{table:notation data}. The health embeddings for each urban living circle are represented as:
\begin{eqnarray}
\mathcal{E}=\left\{\vec{e}_{1},\vec{e}_{2}, \ldots, \vec{e}_{n}\right\}, \vec{e} \in \mathbb{R}^{d}, \forall \vec{e} \in \mathcal{E},
\end{eqnarray} 
where $\vec{e}_i \in \mathbb{R}^{d}$ represents the health embedding of the $i$-th urban living circle $c_i$, and $d$ is the embedding size.

\section{Methodology}
~\
\begin{figure*}[!tb]
	\centering
	
        \includegraphics[width=0.95\textwidth]{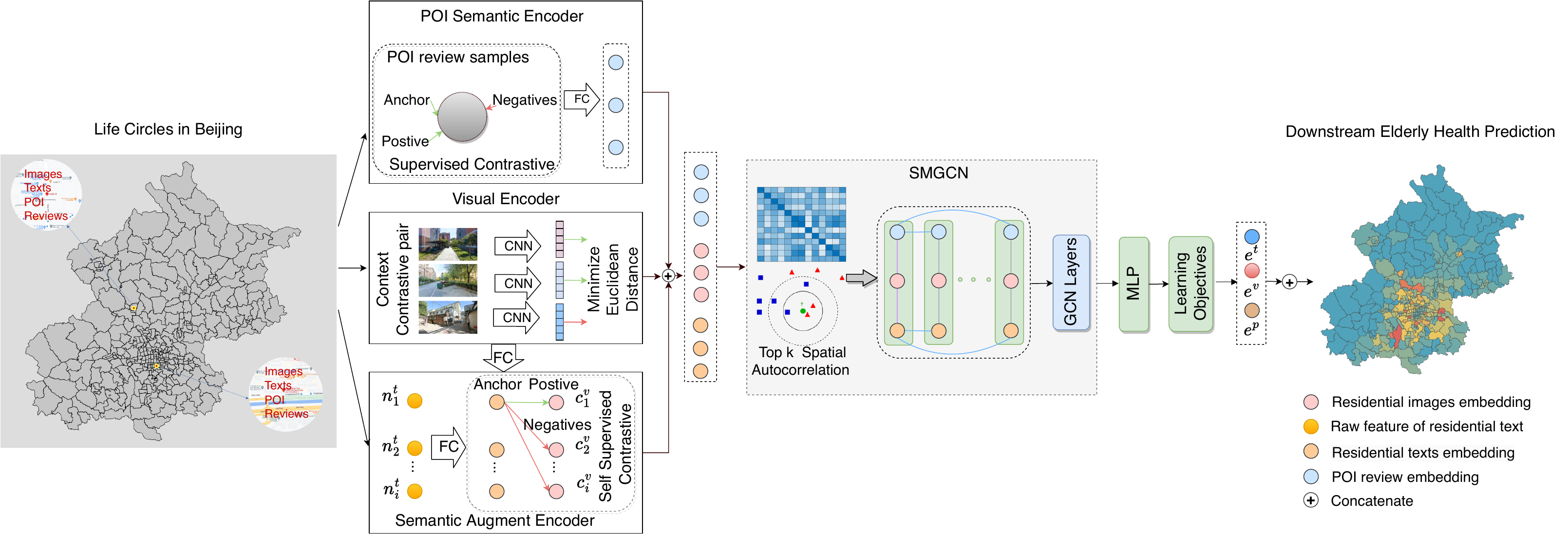}
    \caption{Framework overview of the proposed CureGraph. The visual encoder leverages self-supervised contrastive learning with negative sampling to extract image features. The POI semantic encoder uses supervised contrastive learning to map POI review texts with the same sentiment label closer in the feature space. Additionally, the semantic augment encoder employs cross-modal contrastive learning techniques, integrating visual features to enrich textual semantics. The spatial multimodal graph network (SMGCN) is constructed by incorporating both modality similarity and spatial correlation. Finally, the multimodal embeddings are processed through a multilayer perceptron (MLP) for downstream elderly health prediction tasks. This framework aims to provide robust health predictions by effectively capturing multimodal and spatial interactions.}
	\label{fig::matching}
	\vspace{-2ex}
\end{figure*}

In this section, we present the framework for graph representation learning based on multimodal spatial fusion to generate health embeddings for urban living circles. The proposed framework integrates visual and textual data from residential areas alongside corresponding POI information within each living circle to learn a compact, low-dimensional, and spatially-aware vector representation. This representation is then used to predict the prevalence of four common geriatric diseases among the elderly population in urban areas. An overview of the proposed CureGraph framework is illustrated in Fig.~\ref{fig::matching}.

\subsection{Modality Encoder}
~\
 
To accurately extract features from urban multimodal data, we draw inspiration from the success of contrastive learning in modality encoding. We develop multiple types of contrastive learning encoders to effectively extract features from textual, visual, and POI data related to urban life in residential areas. The core principle of contrastive learning is to achieve both alignment and uniformity~\cite{wang2020understanding}. Specifically, contrastive learning assumes that samples from positive pairs are similar, aiming to maximize the similarity of representations within the same pair while increasing the dissimilarity between representations of different pairs. Consequently, constructing contrastive pairs is a critical step in this approach. In our modality encoder, we leverage both self-similarity and geographical similarity across modalities to construct contrastive pairs. The encoder is designed to map similar data to closely aligned embeddings while maximizing the divergence between embeddings of dissimilar data types.

\subsubsection{Visual Modality Encoder}

Tobler’s First Law of Geography states that objects closer to each other are more likely to be related than those farther apart. Building on this principle, we design a geospatial contrastive learning model to effectively extract image features of urban living circles from residential images. The visual contrastive learning model aligns similar samples by positioning them closer in the projection space while pushing dissimilar samples farther apart. In this context, we assume that images from the same residential area share similar geospatial information.  

To capture this, we define the geographic neighbors of a residential area as its "context", determined by the service radius of the living circle. Specifically, for an image $v$, we identify a positive image $v_{p}$ within the context (geospatial neighbors) and a negative image $v_{n}$ outside the context. The model is trained on a collection of triplets $(v, v_p, v_n)$ using a geospatial contrastive loss function:  
\begin{eqnarray}
\mathcal{L}_{v}^{geo}\left(v, v_{p}, v_{n}\right)=\left[m+\Vert\boldsymbol{x}-\boldsymbol{x_{p}}\Vert_{2}-\Vert\boldsymbol{x}-\boldsymbol{x_{n}}\Vert_{2}\right]_{+} ,
\end{eqnarray}
where $[\cdot]_+$ is the rectifier function, and $\Vert \cdot \Vert_2$ denotes the Euclidean distance. The margin $m$ ensures that the difference between these distances does not grow infinitely large. The embeddings of images $v$, $v_{p}$, and $v_{n}$ are represented as $\boldsymbol{x}$, $\boldsymbol{x_{p}}$, and $\boldsymbol{x_{n}}$, respectively. These embeddings are derived by encoding the images using a pre-trained InceptionV3~\cite{szegedy2016rethinking} model with an embedding dimension of 768.

We also design a contrastive learning framework based on data augmentation to enhance the robustness of feature extraction. The principle of \textit{self-similarity} ensures that augmented data variants remain similar to the original samples. In computer vision~\cite{jaiswal2020survey}, contrastive samples of an image can be generated through data augmentation techniques such as rotation, cropping, masking, and adding noise. For this study, we use rotation and flipping to create two augmented images, $v$ and $v^+$, for each original image. Let $\boldsymbol{x}$ and $\boldsymbol{x}^+$ represent their embeddings, and consider other images in the same batch as negative samples. The loss function for a mini-batch of $N$ pairs is expressed as:
\begin{eqnarray}
\mathcal{L}_{v}^{aug}=-\log\frac{\exp \left(\operatorname{sim}\left(\boldsymbol{x}_{v,i}, \boldsymbol{x}_{v,i}^+\right)/\tau\right)}{\sum_{j=1}^{N}\exp \left(\operatorname{sim}\left(\boldsymbol{x}_{v,i}, \boldsymbol{x}_{v,j}^+\right)/\tau\right)} ,
\end{eqnarray}
where $\operatorname{sim}(\cdot)$ denotes the cosine similarity between two vectors, and $\tau \in \mathbb{R}$ is a temperature hyperparameter that controls the concentration of the distribution. 

Finally, the overall contrastive loss for the visual modality encoder, $f_v(\cdot)$, combines geospatial and augmentation-based components:
\begin{eqnarray}
\mathcal{L}_{v}= \mathcal{L}_{v}^{geo} + \mathcal{L}_{v}^{aug}.
\end{eqnarray}

After obtaining the feature representation for each image through the encoder $f_v(\cdot)$, we apply an averaging operation to derive the corresponding visual feature for each urban living circle $c_i$. This is expressed as:
\begin{eqnarray}
\boldsymbol c_{i}^{v}=\frac{1}{M_{i}}\sum_{j=1}^{M_{i}} \boldsymbol x_{v,ij} ,
\end{eqnarray}
where $M_i$ denotes the number of images in the living circle $c_i$, and $\boldsymbol{x}_{v,ij}$ represents the raw feature representation of the $j$-th image extracted by the visual modality encoder.

\subsubsection{Textual Augmentation Encoder}

In the field of computer vision (CV), contrastive learning has proven to be a powerful method for feature extraction. Similarly, in natural language processing (NLP), contrastive learning is gaining traction due to its ability to learn robust representations using contrastive pairs that pull similar samples closer while pushing dissimilar ones apart~\cite{wang2020understanding}. A critical aspect of contrastive learning lies in constructing positive and negative samples, with various text augmentation techniques—such as adversarial attacks, sentence shuffling, cutoff, and dropout—being widely adopted~\cite{yan2021consert}.

Building on this principle, we leverage geographical similarity between residential texts and images that share the same spatial information. Our fundamental assumption is that texts and images in close geographical proximity exhibit similar semantic spatial representations. Notably, textual descriptions of residential areas within living circles often consist of long sentences. To capture the full semantic modality of text, we draw inspiration from cross-modal contrastive learning, aligning textual representations with their visual counterparts.

For the text modality encoder, we introduce a visually-enhanced text semantic augmentation model based on cross-modal contrastive learning. The core insight is that the semantic representation of a region should closely align with its visual features. Specifically, we use the pre-trained language model \texttt{hfl/chinese-bert-wwm-ext}, following~\cite{9599397}, to encode the textual modality. A fully connected network then projects the representations of both textual and visual modalities into a shared latent space for contrastive learning. The outputs for textual and visual modalities of $c_i \in \mathcal{C}$ are expressed as:  
\begin{equation} 
\label{eqn2}
\begin{aligned} 
&h_{t,i} = W_e^{t}\boldsymbol{n}_i^{t} + b_i^t ,& \\
&h_{v,i} = W_e^{v}\boldsymbol{c}_i^{v} +b_i^v ,& 
\end{aligned}
\end{equation}
where $\boldsymbol{n}_i^{t}$ represents the raw feature vector of the textual modality, and $\boldsymbol{c}_i^{v}$ is the visual embedding of living circle $c_i$. $h_{t,i}$ and $h_{v,i}$ denote the hidden features of the textual and visual modalities, respectively.

To align textual and visual modalities, we extend the traditional InfoNCE loss for the textual augmentation encoder. The contrastive loss is defined as:  
\begin{eqnarray}
\mathcal{L}_{t}^{aug}=-\log\frac{\exp \left(\operatorname{sim}\left(\boldsymbol{h}_{t,i}, \boldsymbol{h}_{v,i}\right)/\tau\right)}{\sum_{j=1}^{N}\exp \left(\operatorname{sim}\left(\boldsymbol{h}_{t,i}, \boldsymbol{h}_{v,j}\right)/\tau\right)} ,
\end{eqnarray}
where the loss is computed over a mini-batch of $N$ samples. 

Unlike previous studies on urban region representation that rely on single or multi-view-based contrastive loss within the same modality or employ standard data augmentation techniques, our proposed cross-modal contrastive loss integrates visual information captured in living circles into textual representations. This innovative approach leverages the complementary nature of visual and textual data, enhancing the overall understanding of urban regions and enabling more accurate and comprehensive representations.
\subsubsection{POI Textual Modality Encoder}

We employ a combination of geographic similarity and data augmentation techniques to train the POI textual modality encoder. Unlike the unsupervised learning approach in the textual augmentation encoder, this encoder utilizes a supervised contrastive learning model. To ensure semantic accuracy and effectively evaluate the quality of services within a community, we incorporate the sentiment tone and category of POI review texts within the living circle. The primary objective of the POI textual modality encoder is to group reviews with similar sentiment tones while distinguishing those with differing sentiments.  

Specifically, we map relevant POIs to the designated living circle, guided by the service radius of the residential area. In order to ensure proximity among POI review text embeddings within the same sentiment category, we employ POI ratings as labels to categorize review texts accordingly. For each POI review text $p$ located within the living circle, we utilize the dropout technique twice as a method of data augmentation. This strategy aims to enrich the dataset with additional positive examples, thereby facilitating the generation of a greater number of positive sample pairs. In the field of NLP, the essence of contrastive learning techniques lies in the creation of positive pairs characterized by sufficient variability. This variability is crucial for training models to discern distinct text representations. Unlike traditional data augmentation methods, which directly alter text content (e.g., through cropping or word deletion) and may inadvertently modify the original intent or introduce noise, our approach leverages the BERT model. By applying diverse dropout masks twice, we generate two unique representations for the same sentence without altering the text directly. This method not only maintains the original text's integrity but also exploits the model's inherent randomness to enhance the diversity between representations.

The goal of our model is to increase the similarity scores among samples within the same category while reducing similarity scores for samples from different categories. Let $\boldsymbol{x}_{p,i}$ and $\boldsymbol{\tilde{x}}_{p,i}$ represent the textual modality embeddings of POI reviews $p$ and $\tilde{p}$, respectively, where $\tilde{p}$ shares the same label as $p$. The sentiment-based, geographically supervised contrastive loss is computed as:
\begin{equation}
\mathcal{L}_{p} = \sum\limits_{i\in I}\frac{-1}{|P(i)|}\sum\limits_{p\in P(i)}\log\frac{\exp ({\mathrm{sim}}(\boldsymbol{x}_{p,i}, \boldsymbol{\tilde{x}}_{p,i})/\tau)}{\sum_{j\in A(i)}\exp (\mathrm{sim}(\boldsymbol{x}_{p,i}, \boldsymbol{\tilde{x}}_{p,j})/\tau)} ,
\end{equation}
where $P(i) \equiv \left\{p \in A(i): \tilde{\boldsymbol{y}}_{p}=\tilde{\boldsymbol{y}}_{i}\right\}$ represents the indices of all positive samples in the mini-batch distinct from $i$, and $|P(i)|$ is its cardinality. $A(i) \equiv I \setminus i$ denotes all indices except $i$. The textual raw feature $\boldsymbol{x}_{p}$ is extracted using the pre-trained language model \texttt{hfl/chinese-bert-wwm-ext}, as described in~\cite{9599397}, with a feature dimension of 768.

To represent each POI within a living circle, we extract semantic features from the review texts and calculate the average sentiment embedding. Given that POI categories are highly correlated with the facility distribution within a living circle, we use the same pre-trained language model to encode the POI categories. The two components—POI review embeddings and category embeddings—are concatenated to form the final POI textual feature. Subsequently, a fully connected network projects the POI textual modality features into the shared latent space for graph learning. Finally, we compute the aggregated POI textual representation $\boldsymbol{c}_{i}^{p}$ for living circle $c_i$ as:  
\begin{equation} 
\begin{aligned} 
&c_{i}^{p}=\sum_{i=1}^{G_{i}}\sum_{j=1}^{m_{j}}{\frac{1}{m_{j}}}{\boldsymbol x_{p,ij}} ,\\
&h_{p,i} = W_e^{p}\boldsymbol{c}_i^{p} +b_i^p, &
\end{aligned}
\end{equation}
where ${G_{i}}$ is the number of POI categories within each living circle, $m_{j}$ is the number of comments for each POI within the living circle, and $\boldsymbol x_{p,ij} = [\boldsymbol x_{p,ij}^{rv},\boldsymbol x_{p,ij}^{cg}]$. Here, $\boldsymbol x_{p,ij}^{rv}$ is the raw feature representation of the POI reviews extracted by the POI textual modality encoder, and $\boldsymbol x_{p,ij}^{cg}$ is the category representation of the corresponding POI.

\subsection{Extracting Spatial Autocorrelation}\label{subsec:spa_corr} 
~\

Urban data are often influenced by geospatial interactions and diffusion, leading to potential correlations and dependencies among data points. This spatial correlation, usually represented by the spatial weight matrix, quantifies the degree of interdependence between data at one location and data at other locations. In this study, spatial autocorrelation coefficients capture both geographical distance and POI-related information within the community living circles. The spatial autocorrelation matrix $\mathcal{H}$ is used to measure the pairwise similarity between each pair of living circles. By incorporating spatial autocorrelation into the graph network, we enhance the learning of modal representations within the community.

To calculate the spatial autocorrelation, we first determine the normalized geographical distance $D_{i,j}$ between the $i$-th living circle $c_i$ and the $j$-th living circle $c_j$. Next, we use the TF-IDF model to assess the importance of each POI category within a living circle, treating POIs as words and a region as a document that describes the functional distribution of the living circle. The urban function similarity between circles is then computed as ${F}_{i,j} = \operatorname{sim}({c_i}, {c_j})$.  

Finally, the combined spatial autocorrelation coefficient is expressed as:  
\begin{eqnarray}
S_{i,j}= \frac{\operatorname{sim}({c_i}, {c_j})}{\log(D_{i,j} + 1)} ,
\end{eqnarray}
where $S_{i,j}$ denotes the spatial autocorrelation coefficient between $c_i$ and $c_j$. This approach enables the proposed model to learn embeddings for multi-modal data in living circles while accounting for inter-region autocorrelations.

\subsection{Graph Construction}
~\

In this section, we present our approach to modeling and constructing the relationships between nodes in community living circles. As defined in Section \ref{sec:prob_state}, we characterize urban community living circles using a 1000-meter service radius around residential areas. Each community living circle encompasses information such as texts, images, and POIs related to the corresponding residential areas. Based on the modal encoder, we can obtain representations of these multimodal data within each community living circle. Specifically, at the spatial street level, we assume that the streets with $n$ residential areas can be represented as an undirected graph ${G}=({V}, {E}, {N})$, where ${V}=\left\{v_{i}\right\}_{i=1}^{3n}$ denotes the three feature nodes in the urban living circle, ${E}$ denotes the edges connecting nodes, and ${N}$ represents the neighborhood of each node. We construct the graph as follows:

\begin{itemize}
    \item \textit{\textbf{Nodes.}}
Each living circle encompasses a residential area, represented by three nodes initialized with the textual modality $\boldsymbol c^{t}$, visual modality $\boldsymbol c^{v}$, and POI textual modality $\boldsymbol c^{p}$, respectively. In accordance with the administrative division at the street level, we construct a graph with $3n$ nodes based on the $n$ living circles in the street.

\item \textit{\textbf{Edges and Edge Weighting.}}
Given the high spatial autocorrelation within the modal information of community living circles, we assume that each node connects to others within the same community living circle. For example, $\boldsymbol{c}^{t}$ will be connected to both $\boldsymbol{c}^{v}$ and $\boldsymbol{c}^{p}$ within the graph. Additionally, for any two nodes of the same modality in different living circles, we connect the top-k nearest nodes based on geographic proximity and modal similarity. Consequently, the graph contains two types of edges: one type connects nodes of different modalities within the same living circle (intra-community edges), and the other connects nodes of the same modality (inter-community edges). We apply different edge-weighting strategies for these two types of edges.

\end{itemize}

The edge weight increases as node similarity increases, reflecting the stronger information interaction between similar nodes. For intra-community edges, we capture the similarity between node representations. Following the approach in~\cite{skianis2018fusing}, we use angular similarity to calculate the edge weight between nodes connected in this manner. The edge weight is computed as:
\begin{eqnarray}
{W}_{ij}^1 =1-\frac{\arccos\left(\operatorname{sim}\left(\boldsymbol x_{i}, \boldsymbol x_{j}\right)\right)}{\pi} ,
\end{eqnarray}
where $\boldsymbol{x}_i$ and $\boldsymbol{x}_j$ are the feature representations of the $i$-th and $j$-th nodes, and $\operatorname{sim}(\cdot)$ denotes cosine similarity.

For the second type of edge, we adhere to the spatial autocorrelation framework described in Subsection \ref{subsec:spa_corr}, employing a top-k locality principle. This ensures that for each urban living circle $c_i$, we connect only the top-k nearest modal nodes within the living circle. The edge weights are determined by both the geographic proximity and modality similarity of the corresponding residential areas. Nodes that are closer in space within a living circle are assigned higher edge weights, ensuring stronger connections for geographically proximate nodes. The edge weight is calculated as:
\begin{eqnarray}
{W}_{ij}^2 = \frac{{W}_{ij}^1}{\log(D_{i,j} + 1)} , \quad c_{j}\in{C}_{i}^{Top-K},
\end{eqnarray}
where ${C}_{i}^{Top-K}$ represents the top-k most correlated candidate regions for the living circle $c_i$ based on spatial autocorrelation, and $D_{i,j}$ is the Euclidean distance between nodes $i$ and $j$. In CureGraph, we combine the top-k locality concept with modality representation similarity to optimize the calculation and search for parameters.

Our proposed composition method simulates the intra-region connections within the living circle modals while comprehensively capturing the inter-region correlations in the modality space. This innovative approach enables a more comprehensive understanding of the relationships and dependencies among different modalities, resulting in more accurate and robust representations of urban regions. By incorporating both intra-region and inter-region correlations, we can effectively capture the fundamental structure and inherent patterns in the data, thus boosting the overall performance of our model.

\subsection{Base Model}

~\

We employ a deep GCN as the base model to develop SMGCN, which learns efficient node representations in a graph. By modeling the spatial relationships in a graph’s multimodal context, SMGCN captures local structural information and the relationships between nodes, enabling it to generate feature representations for each node. Additionally, SMGCN further enhances these representations by integrating information from neighborhoods across different modalities, improving the understanding of graph structure and inter-node relationships while strengthening the encoding of contextual dependencies.

Given an undirected graph ${G}=({V}, {E}, {N})$ and the input vertex features $[\mathcal{T},\mathcal{V},\mathcal{P}]$, we approximate first-order spectral graph convolutions by propagating input data through augmented adjacency and degree matrices. To learn distributed representations of each node in the graph, we adopt a layer-wise propagation rule with residual connections, as proposed in~\cite{chen2020simple}. This update rule for node features is expressed as:
\begin{equation} \label{eqn3}
\begin{aligned} 
  &{H}^{(k)}=\sigma\left(\left((1-\alpha)\tilde{\mathcal{P}}{H}^{(k-1)}+\alpha {H}^{(0)}\right)\left(\left(1-\beta_{k-1}\right) {I}+\beta_{k-1} {W}^{(k-1)}\right)\right),    &\\
  &\beta_{k}=\log\left(\frac{\eta}{k}+1\right),&
\end{aligned}
\end{equation}
where ${H}^{(k)}$ denotes the node feature after the $k$-th aggregation iteration, and $\tilde{\mathcal{P}} = \tilde{\mathcal{D}}^{-1/2} \tilde{\mathcal{A}} \tilde{\mathcal{D}}^{-1/2}$ is the renormalized graph Laplacian matrix of $G$. The parameters $\alpha$, $\beta_{k}$, and $\eta$ are hyperparameters, while $\sigma(\cdot)$ is the activation function (sigmoid), and ${W}^{(k)}$ is a learnable weight matrix. The parameter $\beta_{k}$ ensures that the weight matrix decays adaptively as more layers are stacked. The initial node features, ${H}^{(0)}$, are initialized with $[\tilde{h}_{t,i}, h_{v,i}, h_{p,i}]$, where $\tilde{h}_{t,i}$ is the augmented textual feature from the textual augmentation encoder, and $I$ is the identity mapping matrix.

After $k$ aggregation iterations, we obtain the fused node features for different modalities, denoted as ${e}{t}$, ${e}{v}$, and ${e}_{p}$. These features are then processed by a multi-layer MLP with ReLU activation to transform the concatenated features corresponding to the same living circle. The final representation is used to predict the prevalence of four common geriatric diseases among the elderly. 

The training algorithm for our base model is detailed in Algorithm~\ref{alg:Algorithm}. The final node features are computed as:
\begin{eqnarray}
e = f_\theta(e^{*}) ,
\end{eqnarray}
where $e^{*} = [{e}_{t}, {e}_{v}, {e}_{p}]$, and $f_\theta(\cdot)$ is an MLP with four layers, parameterized by $\theta$.

\begin{algorithm}[!t]
\caption{Training Algorithm of SMGCN.}
\label{alg:Algorithm}
\LinesNumbered
    \KwIn{\\
    \quad Multimodal embeddings of the training urban living circles:\\
    ${\mathcal{T}, \mathcal{V}, \mathcal{P}} = \left\{\vec{c}_{1}^{t,v,p}, \vec{c}_{2}^{t,v,p}, \ldots, \vec{c}_{i}^{t,v,p}\right\}$
    
    \quad Spatial correlation of the training urban living circles:\\ $\mathcal{H}=\{\vec{h}_i\mid i=1,2,\ldots,n\}, \vec{h}_i=(c_1,c_2,\ldots,c_{|\vec{h}_i|})$ and $c_{i}$ is the $i$-th  living circle, ${C}_{i}^{Top-K}$ is the top-k most correlated candidate regions for $c_{i}$ based on $\mathcal{H}$.
    }
    \KwOut{\\
    \quad Learned distributed and low dimensional health embeddings $e$ for urban living circle.
    }

        \For {$t \gets 1 \text{ to } Epoch$}{

        %for i in Batchsize: \\
        $h_{v,i} = Linear({c}_{i}^{v})$, $h_{p,i} = Linear({c}_{i}^{p})$ \;
         
        ${W}_{ij}^1 \leftarrow 1-\frac{\arccos\left(\operatorname{sim}\left(\boldsymbol x_{i}, \boldsymbol x_{j}\right)\right)}{\pi}$ , $\boldsymbol x \in \{\mathcal{T}, \mathcal{V}, \mathcal{P}\}$ \;
       
        ${W}_{ij}^2 \leftarrow \frac{{W}_{ij}^1}{\log(D_{i,j} + 1)} ,\quad c_{j}\in{C}_{i}^{Top-K}$\;
        conduct graph ${G}=({V}, {E}, {N})$, initialized with $[\tilde h_{t,i}, h_{v,i}, h_{p,i}]$ \;
    
        $e^{*} = f([{e}_{t}, {e}_{v}, {e}_{p}])$ where $f$ is the base model \;
        $e = f_\theta(e^{*})$ \;
        $ loss \Longleftarrow\mathcal{L}$ where $\mathcal{L}$ is Equation (19)\;
        optimizer.step($loss$) \;}
       
\end{algorithm}

\subsection{Learning Objectives}
~\

Given a multi-modal spatial GCN, the features of each node in the graph are updated through k-layer iterations. The primary objective of CureGraph is to learn effective low-rank vectors in the embedding space after multi-modal fusion, while preserving spatial autocorrelation. Specifically, CureGraph aims to achieve two goals: (1) capture the rich and meaningful semantics within living areas and fuse them effectively, and (2) accurately model the spatial autocorrelation between living circles. Furthermore, CureGraph is based on the assumption that the texts within residential areas should be highly correlated with the street-view and POI information in their spatial context. Therefore, the model aims to minimize the distance between neighborhoods that have both street-view and POI representations in the embedding space. To train the model effectively, we design a task that reconstructs the spatial autocorrelation coefficients based on the corresponding embeddings and minimizes their Euclidean distance. The learning objective is defined as follows:
\begin{eqnarray}
\mathcal{L}=\sum_{i, j}\left(\mathcal{S}_{i, j}-e_{i}^{T}e_{j}\right)^{2} + \lambda\sum_{m\in\{v,p\}}\operatorname{dist}({e}_{t}, {e}_{m}) ,
\end{eqnarray}
where \( \mathcal{S}_{i, j} \) represents the spatial autocorrelation coefficient between living circles \( c_i \) and \( c_j \), and \( \operatorname{dist(\cdot)} \) is the distance function used in the vector space (here, we use Euclidean distance). The hyperparameter \( \lambda \) controls the contribution of the spatial distance loss to the global loss function. The term \( e_m \), for \( m \in \{v, p\} \), refers to the visual and POI textual modality embeddings, respectively.

\section{Experiments} 

\hspace*{\fill}

In this section, we conduct experiments on real-world datasets to evaluate the performance of our method in comparison to various baselines. Additionally, we perform a series of ablation studies across different modality settings to validate the contribution of each modality fusion in our approach. We also analyze the impact of contrastive learning on the multi-modal encoder. Finally, we carry out a neighborhood similarity analysis of the health embeddings for living circles across multiple spatial levels.

\subsection{Experimental Setup}
~\
\subsubsection{Datasets}
\noindent \textit{\textbf{Real Estate Data}}. This dataset is collected from Lianjia\footnote{https://bj.lianjia.com/}, the largest real estate trading platform in China. It contains information on over 8,000 properties in Beijing and over 7,000 in Shanghai. Each property record includes details such as the name of the residential area, street address, latitude and longitude, price, the total number of buildings and units, and corresponding textual descriptions, image links, and quantities. The textual descriptions provide insights into the residential area’s environment, covering aspects such as supporting facilities, transportation, surrounding amenities, and target residents. Each residential area also includes a varying number of images that visually depict leisure facilities, environmental greening, and other local features. Additionally, the dataset is supplemented with information from the 2020 permanent population census\footnote{https://www.stats.gov.cn/} for both Beijing and Shanghai.
~\\
\noindent \textit{\textbf{POI Check-in Data.}}
The POI check-in data consists of 753,891 records from Beijing and 549,817 records from Shanghai. The data is sourced from various platforms, including Dianping\footnote{https://www.dianping.com/}, Meituan\footnote{https://www.meituan.com/}, Baidu Maps\footnote{https://map.baidu.com/}, Ctrip\footnote{https://hotels.ctrip.com/}, Tujia\footnote{https://m.tujia.com/}, and Elong\footnote{https://www.elong.com/}. Dianping is a well-known platform in China for local life services, providing reviews and facilitating transactions. Meituan offers a wide range of services, from dining to entertainment, while Baidu Maps provides travel-related tools such as route planning, navigation, and location queries. Ctrip, Tujia, and Elong are major travel and ticketing platforms where users can leave reviews about their experiences.

\begin{figure*}[hbtp]
        \centering
        \subfigure[Beijing]{
                \includegraphics[width=0.39\linewidth]{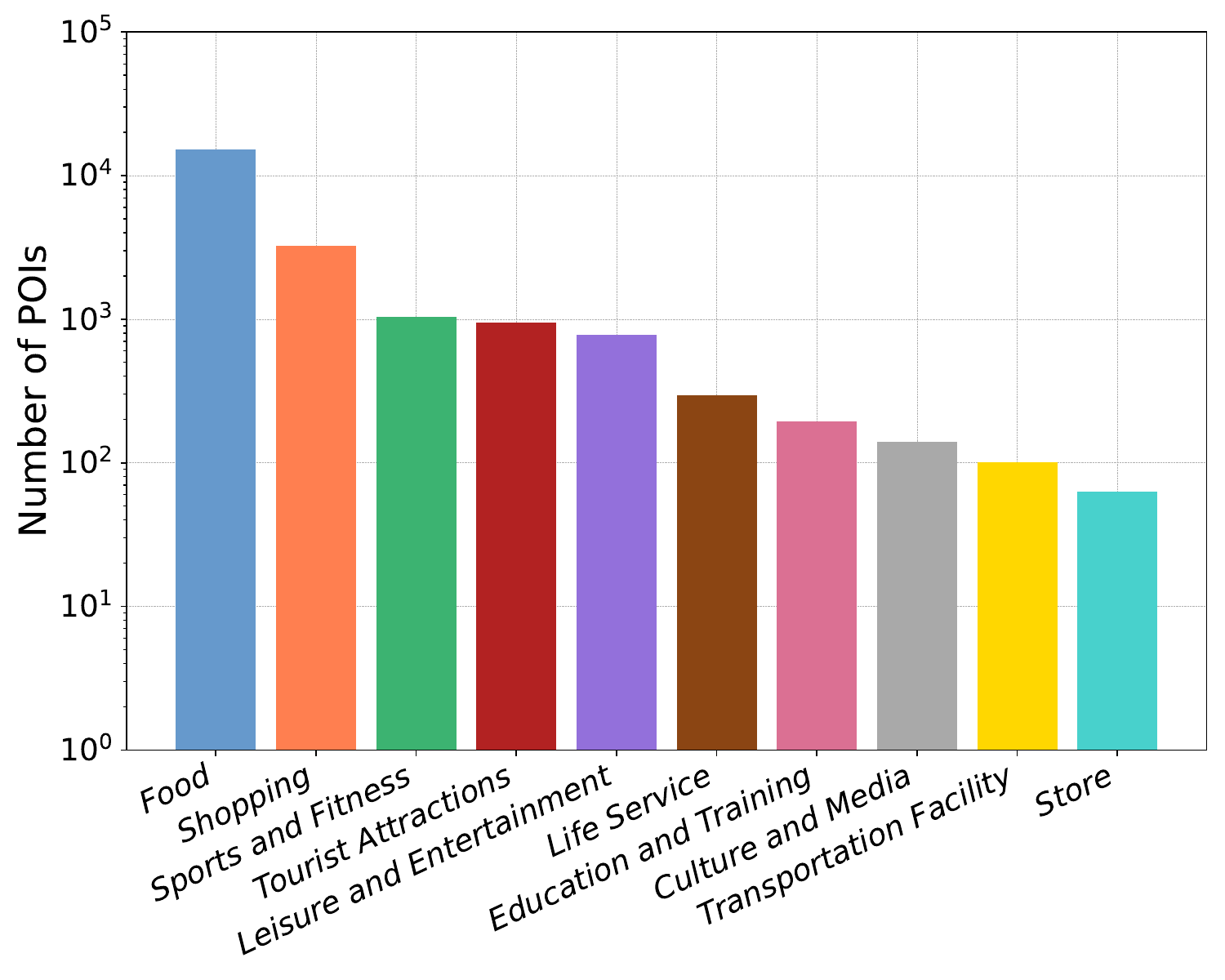}}
        \subfigure[Beijing]{
                \includegraphics[width=0.39\linewidth]{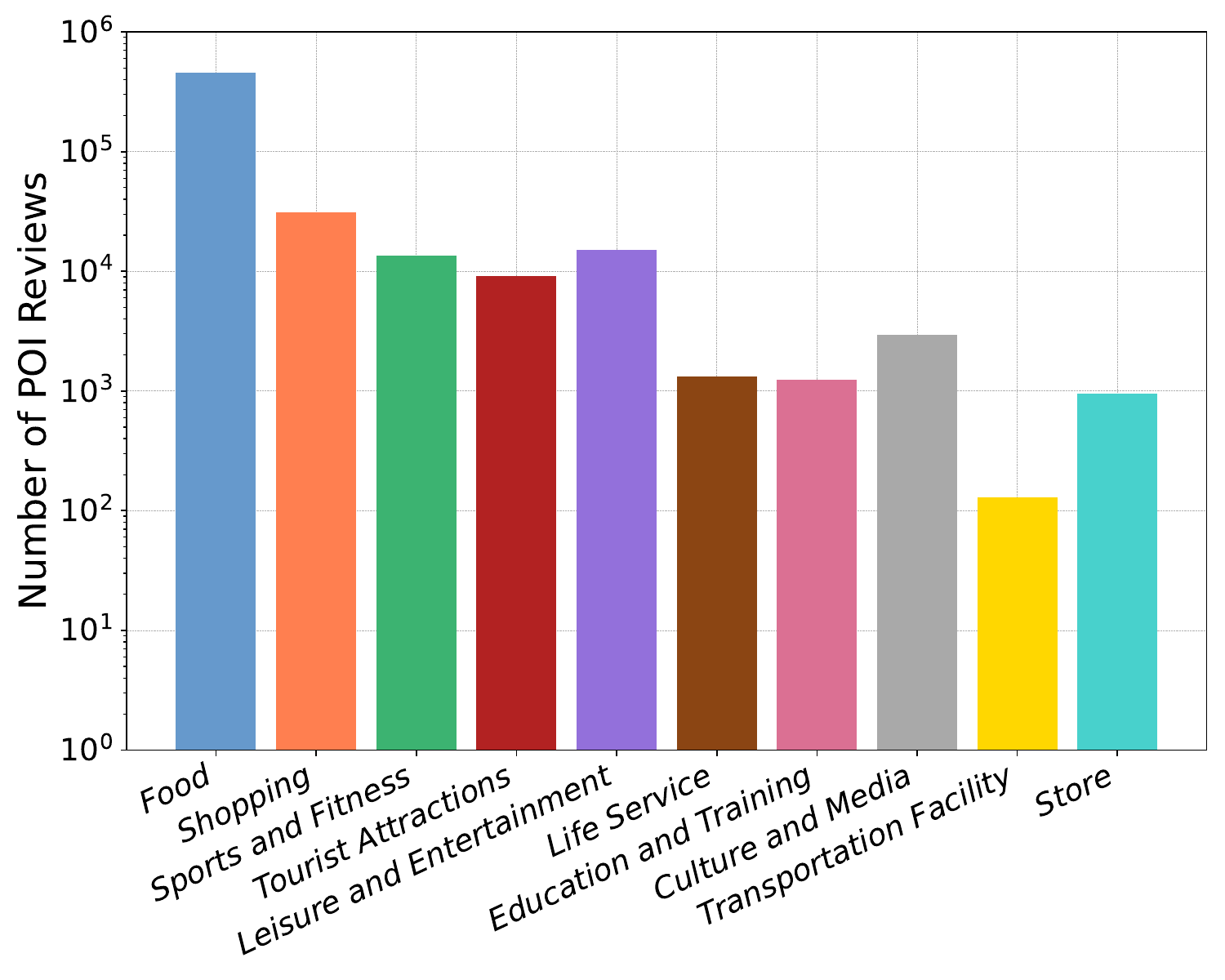}} \\
        \setcounter{subfigure}{0}
        \subfigure[Shanghai]{
                \includegraphics[width=0.39\linewidth]{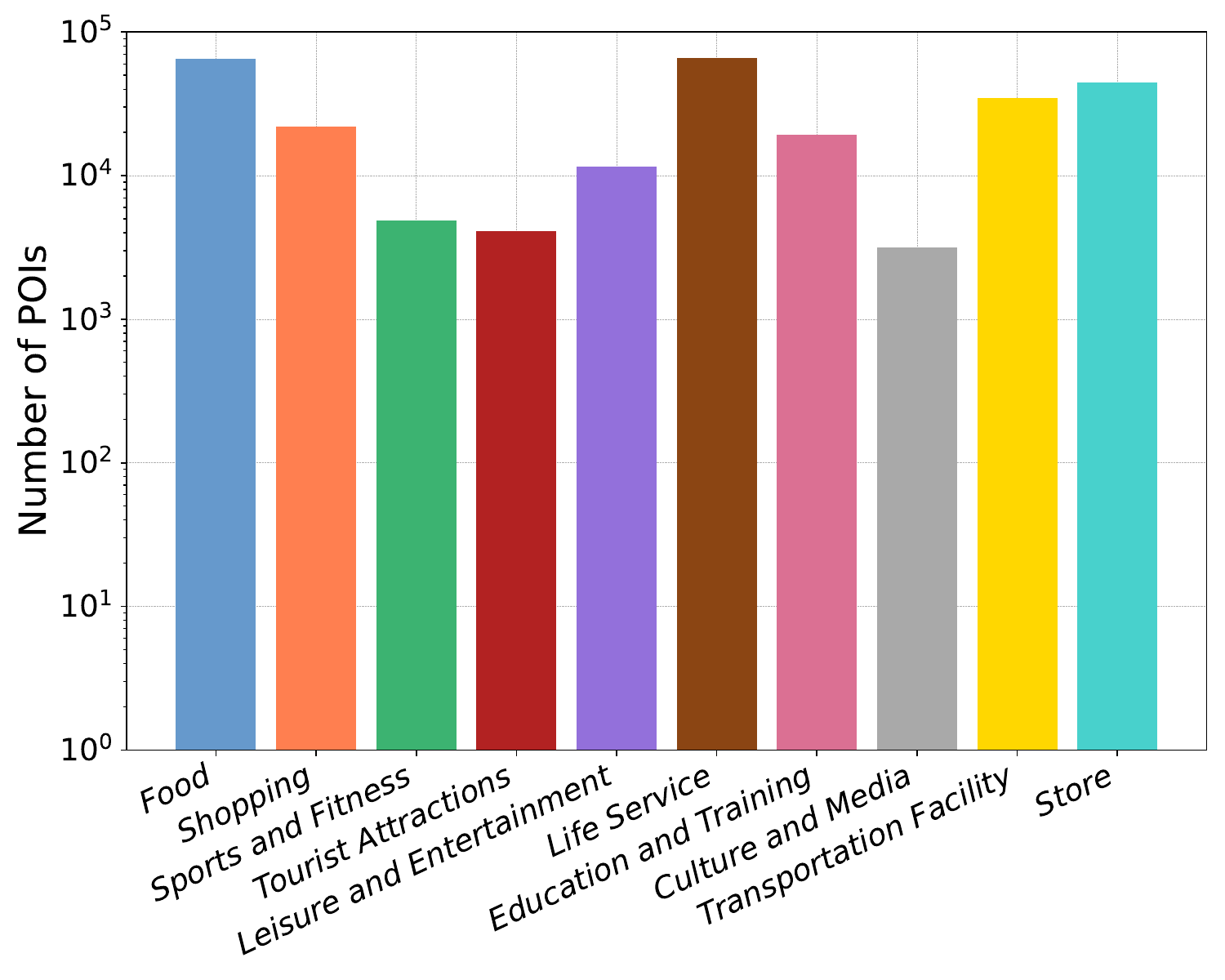}}
        \subfigure[Shanghai]{
                \includegraphics[width=0.39\linewidth]{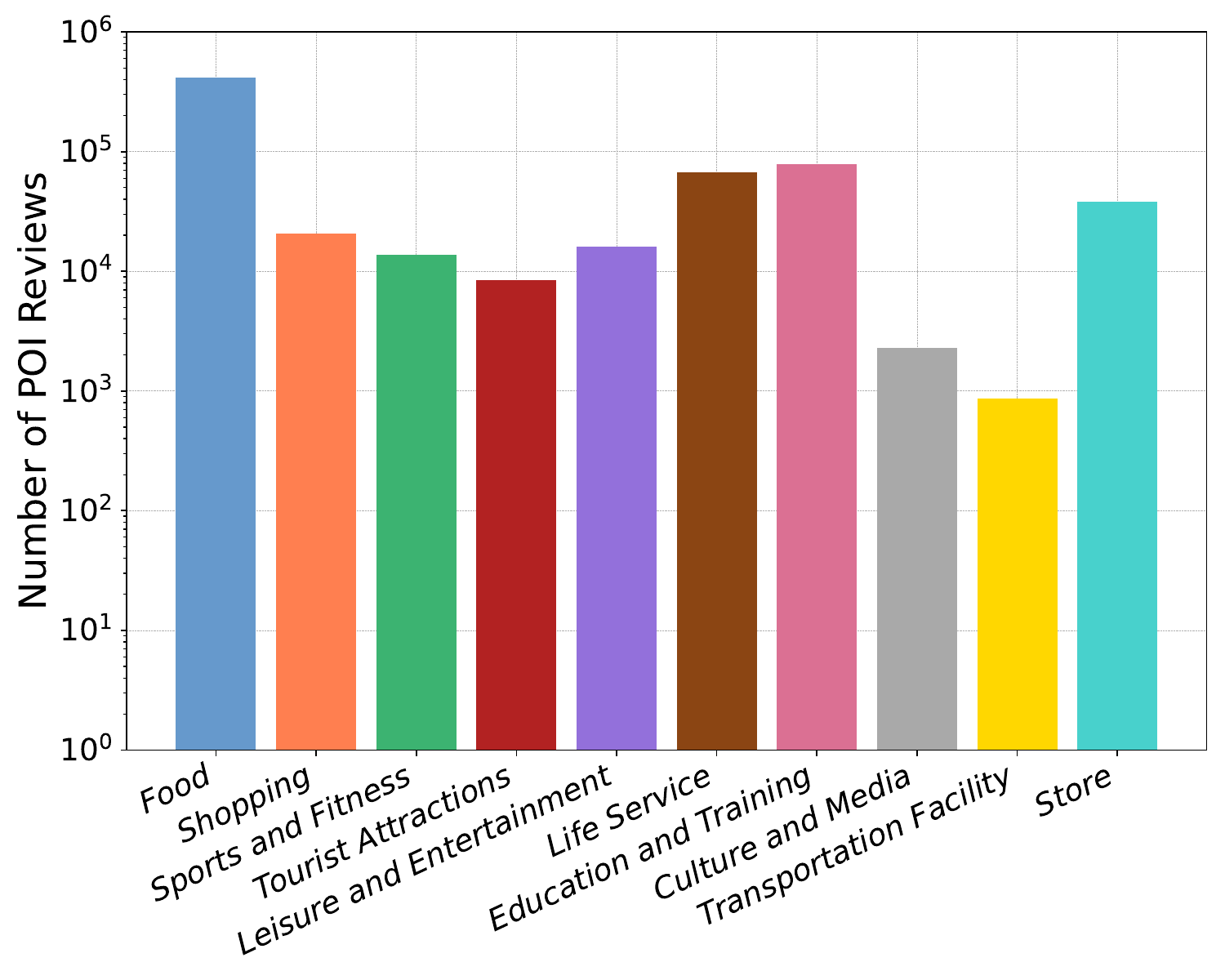}}
        \caption{Distribution of POI categories and the number of reviews across the datasets.}
        \label{fig:distribution}
                %\vspace{-10px}
\end{figure*}

The dataset records user check-ins from January 2019 to December 2019, each associated with a unique user ID, venue ID, user comments and ratings, as well as price and source information. Although longitude and latitude information for POIs are not provided, we are able to match POIs with geographical data using the Baidu Maps API, as all platforms in the dataset utilize Baidu Maps for their services. The POIs are categorized into 10 main types, which are selected based on their relevance to chronic disease factors among the elderly. These categories include food, shopping, sports and fitness, tourist attractions, leisure and entertainment, life services, education and training, culture and media, transportation facilities, and stores. The distribution of POIs and reviews across categories is shown in Fig.~\ref{fig:distribution}.

\noindent \textit{\textbf{Geriatric Disease Data.}}
Due to strict patient privacy protocols and the detailed nature of our research, which focuses on fine-grained residential areas, obtaining comprehensive geriatric disease data presents significant challenges. To address this issue, we use simulated data for tasks related to chronic diseases in the elderly population. We focus on a set of common chronic diseases, including MCI, hypertension, diabetes, and MDD. The prevalence of these diseases is simulated for community residents aged 60 and above in both Beijing and Shanghai. For Beijing, the simulation draws on contrasting morbidity rates between urban core and suburban areas, as reported in medical journals~\cite{tang1991prevalence,ma2015epidemiological,yang2000prevalent,liu2015prevalence}. For Shanghai, the simulation references data from~\cite{ding2015prevalence,dong2017determinants,hu2022prevalence,NKLL201803014}. Additionally, the population of residents aged 60 and above is simulated based on the average household size in each district, as reported in the 2020 population statistics for Beijing and Shanghai, along with the total number of housing units listed on the Lianjia website.

A series of preprocessing steps are implemented to enhance the quality and clarity of the dataset for subsequent tasks. First, to ensure comprehensive multimodal information collection in residential areas, we set a 1000-meter service radius for each area. Next, POI rating samples with a score of 0 within the defined living circle are excluded to ensure compatibility with the text content. Due to the large size of the POI check-in dataset, only a 10\% subset of POI review samples from each category is used during the training of the modal encoder. These samples are then further split into testing and validation sets at a 7:3 ratio. The detailed specifications of our datasets are provided in Table~\ref{table:dataset}.

\begin{table}[!t]
\centering
\caption{Statistics of the experimental datasets for Beijing and Shanghai.}

\label{table:dataset}
\resizebox{\linewidth}{!}{
\begin{tabular}{l|p{0.45\linewidth}|p{0.45\linewidth}}
\toprule
\multirow{2}{*}{\textbf{Dataset}}
& \multicolumn{2}{c}{\textbf{Description}} \\
\cmidrule{2-3}
& \textbf{Beijing} & \textbf{Shanghai} \\
\midrule
Streets & 204 districts divided by administrative streets. & 191 districts divided by administrative streets. \\
\midrule
Living circles & 5087 living circles divided by the 1km radius of residential areas. & 3997 living circles divided by the 1km radius of residential areas.\\
\midrule
Real estate data & 5087 residential area texts and around 45 thousand community photos in the living circles. & 3997 residential area texts and around 38 thousand community photos in the living circles. \\
\midrule
POI check-in data & Over 700 thousand check-in reviews from more than 40 thousand POIs with 10 major categories, along with 143,499 POI locations with 210 middle categories from Baidu Maps. & Over 500 thousand check-in reviews from more than 37 thousand POIs with 10 major categories, along with 275,827 POI locations with 98 middle categories from
Baidu Maps.\\
\midrule
Geriatric disease data & \multicolumn{2}{l}{Prevalence of MCI~\cite{tang1991prevalence,ding2015prevalence}; Prevalence of hypertension~\cite{ma2015epidemiological,dong2017determinants}; Prevalence of diabetes~\cite{yang2000prevalent,hu2022prevalence}; Prevalence of MDD~\cite{liu2015prevalence,NKLL201803014}.}  \\

\bottomrule
\end{tabular}
}
\end{table}

\subsubsection{Baselines}
To demonstrate the performance of the proposed method, we compare CureGraph against the following algorithms:

\begin{itemize}

\item \textbf{LINE}~\cite{tang2015line}. This method optimizes an objective function designed to preserve both local and global network structures using an edge-sampling algorithm. In this study, we apply the LINE technique to multi-modal graphs of living circles and concatenate the embeddings of each modality.

\item \textbf{Node2Vec}~\cite{grover2016node2vec}. Node2Vec employs biased random walks to learn latent features that maximize the likelihood of preserving local network structures. Here, we use Node2Vec on multi-modal graphs of living circles and concatenate the embeddings of each modality.

\item \textbf{GAT}~\cite{velickovic2017graph}. The Graph Attention Network (GAT) updates the representation of each node by propagating information from its neighbors via an attention mechanism. In this work, we employ a multi-head GAT to learn node representations.

\item \textbf{MVGJR}~\cite{zhang2021multi}. This unsupervised multi-view joint learning approach is tailored for urban region embedding. It leverages the regional correlations of human mobility and inherent regional properties by employing a cross-view attention mechanism.

\item \textbf{Urban2Vec}~\cite{wang2020urban2vec}. This is an unsupervised model for neighborhood representation learning. Urban2Vec integrates information from street views and POI data and assumes that neighborhoods with smaller geographic distances are more likely to share common semantics than those farther apart.

\item \textbf{MuseCL}~\cite{yong2024musecl}. This multi-semantic contrastive learning framework utilizes similarities in human mobility patterns and POI distributions within regions to extract semantic features from visual modalities. These visual features are then integrated with textual data from embedded POI texts using a cross-modality attention mechanism and contrastive learning for region representation learning.
\end{itemize}

\subsubsection{Evaluation Metrics}

To evaluate the performance of the models, we employ three metrics: Mean Absolute Error (MAE) and Root Mean Square Error (RMSE) to quantify prediction errors, and the coefficient of determination ($R^2$) to assess the goodness of fit. These metrics are defined as follows:
\begin{eqnarray}
\text{MAE} = \frac{1}{N}\sum_{i=1}^N|\hat{Y}_i - Y_i|, \quad \text{RMSE} = \sqrt{\frac{1}{N}\sum_{i=1}^N(\hat{Y}_i - Y_i)^2}, \quad R^2 = 1 - \frac{\sum_{i=1}^{N} (Y_i - \hat{Y}_i)^2}{\sum_{i=1}^{N} (Y_i - \bar{Y})^2},
\end{eqnarray}
where $N$ is the total number of instances, $\hat{Y}_i$ denotes the predicted value, $Y_i$ represents the ground truth, and $\bar{Y}$ is the mean of the observed values. To ensure robust and reliable results, all metrics are calculated using $K$-Fold cross-validation, with $K=5$.

\subsubsection{Parameters Settings}

In this section, we detail the experimental settings. For the visual modality encoder, image pairs are randomly selected within the 15-minute living circle for training. The encoder is initialized with parameters pre-trained on ImageNet, with a learning rate of $5 \times 10^{-3}$ and a temperature hyperparameter $\tau$ set to 0.05. For the textual modality encoder, we use BERT as the backbone, pre-trained on a large corpus. The learning rates for the textual augmentation encoder and POI textual modal encoder are set to $1 \times 10^{-5}$, with $\tau$ set to 1 and 0.005, respectively. The GCN has three layers, a learning rate of $5 \times 10^{-4}$, a dropout rate of 0.3, and $K$ set to 20 for the Top-$K$ nearest neighbors. The hyperparameters $\alpha$, $\eta$, and $\lambda$ are set to 0.2, 0.5, and 0.1, respectively. The batch size is fixed at 32 across all modules. For the Urban2vec model, the embedding size is 200, which is reduced to 128 using PCA, as recommended by the authors. Similarly, the embedding size for LINE, Node2Vec, GAT, MVGJR, and our proposed model and its variants is set to 128.

The implementation is based on PyTorch (version 1.11.0)~\cite{paszke2019pytorch}, and all experiments are conducted on an NVIDIA A40 GPU (1.38 GHz, 48 GB GDDR6). We use the Adam optimizer~\cite{kingma2014adam} with a weight decay of $3 \times 10^{-3}$ and train the backbone model for 60 epochs.

\begin{table*}[!t]
    \caption{Prediction results for community elder health outcomes in Beijing across different models.}
    \label{table:results_bj}
    \resizebox{\textwidth}{!}{   
    \begin{tabular}{llllllllllllll}
    \toprule
    \multirow{2}{*}{Model} & \multicolumn{3}{c}{MCI}       & \multicolumn{3}{c}{Hypertension}    &  \multicolumn{3}{c}{Diabetes}   &   \multicolumn{3}{c}{MDD} \\ 
    \cmidrule[0.8pt](lr){2-4}\cmidrule[0.8pt](lr){5-7}\cmidrule[0.8pt](lr){8-10}\cmidrule[0.8pt](lr){11-13}
        & MAE$\downarrow$       & RMSE$\downarrow$     & $R^{2}\uparrow$
        & MAE$\downarrow$       & RMSE$\downarrow$     & $R^{2}\uparrow$
        & MAE$\downarrow$       & RMSE$\downarrow$     & $R^{2}\uparrow$
        & MAE$\downarrow$       & RMSE$\downarrow$     & $R^{2}\uparrow$
    \\ \midrule 
    Node2Vec      & 23.471     & 34.277      & -0.133      
                  & 28.534     & 41.736      & -0.106    
                  & 54.948     & 82.913      & -0.144
                  & 17.412     & 25.007      & -0.127
    \\
    LINE          & 26.689     & 37.571      & -0.361    
                  & 31.745     & 45.241      & -0.299
                  & 62.904     & 89.799      & -0.315
                  & 19.611     & 27.505      & -0.363
    \\ 
    GAT           & 21.600     & 32.376      & -0.010       
                  & 27.017     & 39.836      & -0.007
                  & 51.383     & 77.767      & -0.007
                  & 15.996     & 23.620      & -0.005
    \\
    MVGJR         & 21.243     & 32.072      & 0.003  
                  & 26.825     & 39.420      & 0.002
                  & 51.075     & 77.544      & -0.002
                  & 15.418     & 23.610      & -0.001
    \\
    Urban2vec     & 20.739     & 31.875      & 0.044   
                  & 26.019     & 39.091      & 0.040   
                  & 50.608     & 77.303      & 0.026
                  & 15.608     & 23.370      & 0.033
    \\ 
    MuseCL & \underline{20.473} & \underline{31.489} & \underline{0.052} 
    & \underline{25.714} & \underline{38.780} & \underline{0.045}  
    & \underline{50.197} &   \underline{76.535} & \underline{0.029} 
    & \underline{15.396} & \underline{23.093} & \underline{0.039} 
                     
    \\ \bottomrule
    w/o Top-k sc  & 20.251     & 31.237      & 0.059 	
                  & 25.298     & 38.380      & 0.065 	
                  & 49.098     & 75.593      & 0.049 	
                  & 15.105     & 22.827      & 0.061 
    \\ 
    Ours          & \textbf{20.082} & \textbf{31.015} & \textbf{0.073} 
                  & \textbf{25.048} & \textbf{38.023} & \textbf{0.082} 
                  & \textbf{48.552} & \textbf{74.920} & \textbf{0.066} 
                  & \textbf{14.926} & \textbf{22.589} & \textbf{0.081}
    \\
    \bottomrule
\end{tabular}}
\end{table*}

\begin{table*}[t]
    \caption{Prediction results for community elder health outcomes in Shanghai across different models.}
    \label{table:results_sh}
    \resizebox{\textwidth}{!}{   
    \begin{tabular}{
        >{\arraybackslash}l
        >{\arraybackslash}l
        >{\arraybackslash}l
        >{\arraybackslash}l
        >{\arraybackslash}l
        >{\arraybackslash}l
        >{\arraybackslash}l
        >{\arraybackslash}l
        >{\arraybackslash}l
        >{\arraybackslash}l
        >{\arraybackslash}l
        >{\arraybackslash}l
        >{\arraybackslash}l
    }
    \toprule
    \multirow{2}{*}{Model} & \multicolumn{3}{l}{MCI}    & \multicolumn{3}{l}{Hypertension}    &  \multicolumn{3}{l}{Diabetes}  &   \multicolumn{3}{l}{MDD} \\ 
    \cmidrule[0.8pt](lr){2-4}\cmidrule[0.8pt](lr){5-7}\cmidrule[0.8pt](lr){8-10}\cmidrule[0.8pt](lr){11-13}
        & MAE$\downarrow$       & RMSE$\downarrow$     & ${R}^{2}\uparrow$
        & MAE$\downarrow$       & RMSE$\downarrow$     & ${R}^{2}\uparrow$
        & MAE$\downarrow$       & RMSE$\downarrow$     & ${R}^{2}\uparrow$
        & MAE$\downarrow$       & RMSE$\downarrow$     & ${R}^{2}\uparrow$
    \\ \midrule 
    Node2Vec      & 42.402     & 57.141      & -0.206      
                  & 153.772     & 185.692      & -0.158    
                  & 68.702     & 74.752      & -0.141
                  & 14.703     & 18.384      & -0.106
    \\
    LINE          & 43.231     & 59.580      & -0.291    
                  & 164.710     & 195.884    & -0.294
                  & 73.171     & 79.897     & -0.250
                  & 15.457     & 19.961     & -0.374
    \\ 
    GAT           & 41.708     & 54.283      & -0.005       
                  & 151.209     & 180.405    & -0.009
                  & 66.171     & 73.039     & -0.008
                  & 13.639     & 17.866     & -0.046
    \\
    MVGJR         & 41.946     & 53.255      & 0.014  
                  & 151.065     & 177.456    & 0.008
                  & 64.275     & 73.043     & 0.007
                  & 13.221     & 17.829     & 0.006
    \\
    Urban2vec     & 39.378     & 52.841      & 0.046   
                  & 145.133     & 170.015    & 0.055   
                  & 62.043     & 71.544     & 0.047
                  & 12.921     & 17.580     & 0.033
    \\
    MuseCL        & \underline{39.067}     & \underline{52.567}      & \underline{0.056}  
                  & \underline{143.080}     & \underline{167.375}    & \underline{0.068}
                  & \underline{61.480}     & \underline{70.967}     & \underline{0.058}
                  & \underline{12.528}     & \underline{17.446}     & \underline{0.058}
    \\\bottomrule
    w/o Top-k sc  & 38.378     & 48.608      & 0.081
                  & 139.653     & 165.866    & 0.072
                  & 60.960     & 69.422     & 0.076
                  & 11.256     & 17.298     & 0.063
    \\
    Ours          & \textbf{37.568}     & \textbf{48.405}      & \textbf{0.090} 
                  & \textbf{138.558}     & \textbf{165.147}    & \textbf{0.085} 
                  & \textbf{60.384}     & \textbf{69.087}     & \textbf{0.082} 
                  & \textbf{10.567}     & \textbf{16.267}     & \textbf{0.078}
    \\
    \bottomrule
\end{tabular}
}
\end{table*}

\subsection{Performance Comparison}
~\
\subsubsection{Overall Comparison}

This study aims to predict the prevalence of four health conditions—MCI, hypertension, diabetes, and MDD—among elderly individuals aged 60 and above residing in specific neighborhoods. To achieve this, we conduct experiments using datasets from two cities, Beijing and Shanghai, with the learnt urban living circle health embeddings serving as input. The proposed method is evaluated against various baseline models, and the results are summarized in Table~\ref{table:results_bj} and Table~\ref{table:results_sh}. In these tables, the best-performing results are highlighted in bold, while the second-best results from baseline models are underlined.

Overall, the results across the four geriatric disease prediction tasks demonstrate the superiority of our proposed CureGraph model. For example, in Beijing and Shanghai, CureGraph achieves improvements of up to 33\% and 22\% in MAE and RMSE, respectively, while enhancing the $R^2$ metric by 25\% to 145\% compared to the best-performing baseline methods.

Among the baselines, MuseCL and Urban2vec demonstrate significant superiority over traditional representation learning methods like Node2Vec and LINE, consistently achieving notable performance boosts. These results suggest that these models are more efficient at capturing the complex associations between multimodal spatial data and the health status of the elderly in urban communities. In contrast, LINE's reliance on first-order proximity information and Node2vec's utilization of random walks may be insufficient in capturing the intricate interplay between different modalities and the complex health outcomes of the elderly population. Urban2vec, while effective, depends on distance similarity between images and the fine-tuning of word embeddings for intra-region POI reviews, which poses challenges in aligning image and word semantics. MuseCL, on the other hand, leverages the distributional similarity of POI types to guide the contrastive pairing of images and text. After performing a simple fusion using linear layers, it compares the fused representation with the semantic information of POI types. However, this contrastive fusion approach may inadvertently overemphasize textual semantics, potentially limiting its ability to capture balanced multimodal interactions. Similarly, MVGJR focuses on region correlations derived from region attributes and multi-view learning of POI attributes and check-in data but neglects multimodal semantic interactions. By contrast, our proposed approach captures both the rich semantic dependencies between intersecting modalities and the spatial autocorrelation connections within urban communities. This comprehensive strategy enables our model to outperform others in predicting community health status.

Additionally, we observe that omitting spatial information (e.g., removing Top-k spatial correlation) significantly reduces performance compared to our full approach. This highlights the importance of considering a community’s health profile in relation to its surrounding neighborhoods. In summary, our research demonstrates the effectiveness of neighborhood embeddings in health outcome prediction and underscores the necessity of integrating spatial autocorrelation information to improve the accuracy and robustness of these predictions.

\subsubsection{Comparison Across Spatial Levels}

Our graph construction approach is based on the division of street areas, resulting in a highly interpretable graph representation. This structure enables operations such as averaging, random walks, and pooling, making it versatile for various analytical tasks. In this study, we derive the graph representation of streets by aggregating node representations at the community circle spatial level within the graph. This approach allows us to evaluate whether the proposed CureGraph performs as effectively at the street level as it does at the community level and to explore the similarities and differences between these spatial resolutions.
The results in Table~\ref{table:streets_level_prediction} and Table~\ref{table:streets_level_prediction_sh} demonstrate the superior performance of our proposed CureGraph model at the street level. Despite the coarser granularity, CureGraph consistently achieves state-of-the-art results, with improvements of up to 27\% and 32\% in MAE and RMSE, respectively, in Beijing and Shanghai, as well as a 5\% to 28\% increase in the $R^2$ metric. Notably, for street-level elder health prediction tasks, particularly for conditions such as MDD, CureGraph not only delivers substantial performance gains but also exhibits strong robustness and reliability.

Node2Vec and LINE, in comparison, yield the lowest predictive metrics, highlighting their limited ability to capture complex spatial relationships. The GAT model performs better, dynamically learning relationships between nodes via attention mechanisms, but it falls short in leveraging edge information and managing high-order features. Similarly, MVGJR and Urban2vec fail to fully integrate complementary data from diverse urban morphology and spatial sources. While MuseCL’s multimodal contrastive fusion enables it to outperform these models across all metrics, CureGraph surpasses them by employing modality-specific encoders, spatial synthesis techniques, and effectively capturing both spatial autocorrelation within neighborhoods and spatial heterogeneity across them.

Furthermore, the $R^2$ metric for elder health prediction highlights CureGraph’s superior fitting capability at the street level compared to the community living circle level. This can be attributed to our graph construction methodology, which accounts for spatial dependencies among various community patterns within neighborhoods. Interestingly, improvements in MAE and RMSE metrics at the street level are more pronounced than at the community level. This suggests that CureGraph exhibits heightened sensitivity to spatial autocorrelation at finer spatial resolutions. However, this sensitivity may also amplify the effects of outliers or anomalous data points, influencing these metrics more significantly at detailed spatial levels.

\begin{table*}[t]
    \centering
    \caption{Elder health prediction results at the street level in Beijing.}
    \label{table:streets_level_prediction}
    \resizebox{\textwidth}{!}{  
    \begin{tabular}{llllllllllllll}
    \toprule
    \multirow{2}{*}{Model} & \multicolumn{3}{l}{MCI}       & \multicolumn{3}{l}{Hypertension}    &  \multicolumn{3}{l}{Diabetes}   &   \multicolumn{3}{l}{MDD} \\ 
    \cmidrule[0.8pt](lr){2-4}\cmidrule[0.8pt](lr){5-7}\cmidrule[0.8pt](lr){8-10}\cmidrule[0.8pt](lr){11-13}
        & MAE$\downarrow$       & RMSE$\downarrow$     & ${R}^{2}\uparrow$
        & MAE$\downarrow$       & RMSE$\downarrow$     & ${R}^{2}\uparrow$
        & MAE$\downarrow$       & RMSE$\downarrow$     & ${R}^{2}\uparrow$
        & MAE$\downarrow$       & RMSE$\downarrow$     & ${R}^{2}\uparrow$
    \\ \midrule
    Node2Vec      & 341.615     & 501.024     & 0.489   
                  & 369.277    & 536.631   & 0.630    
                  & 1043.999    & 1612.160    & 0.183
                  & 227.658    & 342.144    & 0.588
    \\
    LINE          & 485.110     & 787.724    & -0.264   
                  & 544.213     & 883.355    & -0.001
                  & 983.920    & 1454.093    & 0.336
                  & 308.203    & 470.488    & 0.221
    \\ 
    GAT           & 330.634     & 490.023    &  0.508      
                  & 374.276    & 561.635    & 0.430 
                  &  973.993   & 1460.159    & 0.340 
                  &  215.207   & 334.477    & 0.640
    \\
    MVGJR         & 327.103      & 485.723     & 0.532    
                  & 373.212    & 557.356    & 0.471   
                  & 977.925    & 1459.034    & 0.335
                  & 223.207   & 266.154  & 0.651
    \\
    Urban2vec     & 249.158       & 361.763     & 0.693 
                  & 262.233        & 389.343       & 0.783
                  & 767.928        & 1109.036     & 0.565
                  & 151.657        & 231.478       & 0.780
    \\
    MuseCL        & \underline{244.434} & \underline{333.503} & \underline{0.702} 
                  & \underline{254.223} & \underline{364.427} & \underline{0.781} 
                  & \underline{574.526} & \underline{833.201} & \underline{0.781} 
                  & \underline{142.485} & \underline{203.176} & \underline{0.854} 
    \\  
    w/o Top-k sc  & 234.147	& 348.757 	& 0.752 	
                  & 241.200 	  & 353.828 	  & 0.839 	
                  & 563.449 	  & 848.603 	  & 0.774 	
                  & 137.272 	  & 191.290 	  & 0.871 
    \\
    Ours          & \textbf{229.761} & \textbf{346.773} & \textbf{0.755} 
                  & \textbf{239.369} & \textbf{343.751} & \textbf{0.848} 
                  & \textbf{560.743} & \textbf{822.355} & \textbf{0.788} 
                  & \textbf{130.551} & \textbf{180.162} & \textbf{0.886}
    \\ \bottomrule
    \end{tabular}}
\end{table*}

\begin{table*}[h]
    \centering
    \caption{Elder health prediction results at the street level in Shanghai.}
    \label{table:streets_level_prediction_sh}
    \resizebox{\textwidth}{!}{   
    \begin{tabular}{
        >{\arraybackslash}l
        >{\arraybackslash}l
        >{\arraybackslash}l
        >{\arraybackslash}l
        >{\arraybackslash}l
        >{\arraybackslash}l
        >{\arraybackslash}l
        >{\arraybackslash}l
        >{\arraybackslash}l
        >{\arraybackslash}l
        >{\arraybackslash}l
        >{\arraybackslash}l
        >{\arraybackslash}l
    }
    \toprule
    \multirow{2}{*}{Model} & \multicolumn{3}{l}{MCI}       & \multicolumn{3}{l}{Hypertension}   &  \multicolumn{3}{l}{Diabetes}   &   \multicolumn{3}{l}{MDD} \\ 
    \cmidrule[0.8pt](lr){2-4}\cmidrule[0.8pt](lr){5-7}\cmidrule[0.8pt](lr){8-10}\cmidrule[0.8pt](lr){11-13}
        & MAE$\downarrow$       & RMSE$\downarrow$     & ${R}^{2}\uparrow$
        & MAE$\downarrow$       & RMSE$\downarrow$     & ${R}^{2}\uparrow$
        & MAE$\downarrow$       & RMSE$\downarrow$     & ${R}^{2}\uparrow$
        & MAE$\downarrow$       & RMSE$\downarrow$     & ${R}^{2}\uparrow$
    \\ \midrule 
    Node2Vec      & 450.866     & 604.675      & 0.487      
                  & 1299.221    & 1644.190     & 0.305    
                  & 1003.658    & 1685.962     & 0.297
                  & 240.593     & 390.954      & 0.528
    \\
    LINE          & 549.689     & 965.301      & -0.372    
                  & 1499.221    & 1953.103     & 0.112
                  & 1209.286    & 1750.076     & 0.171
                  & 289.476     & 495.108      & 0.321
    \\ 
    GAT           & 440.185     & 580.970      & 0.515       
                  & 1272.914    & 1572.611     & 0.435
                  & 801.433     & 1629.305     & 0.342
                  & 236.891     & 340.864      & 0.588
    \\
    MVGJR         & 436.081     & 572.376      & 0.539  
                  & 1153.373    & 1531.367     & 0.492
                  & 790.271     & 1607.493     & 0.369
                  & 229.492     & 335.456      & 0.591
    \\
    Urban2vec     & 358.976     & 451.309      & 0.685   
                  & 1118.217    & 1438.046     & 0.605   
                  & 650.163     & 1433.858     & 0.503
                  & 156.802     & 320.854      & 0.719
    \\
    MuseCL        & \underline{349.883}     & \underline{430.877}      & \underline{0.694}  
                  & \underline{1036.118}   & \underline{1259.877}    & \underline{0.682}
                  & \underline{642.501}    & \underline{1338.832}    & \underline{0.602}
                  & \underline{146.594}    & \underline{302.387}     & \underline{0.726}
    \\  \midrule
    w/o Top-k sc  & 338.083     & 415.842      & 0.738
                  & 966.816     & 1043.490     & 0.781
                  & 579.419     & 1112.819     & 0.689
                  & 138.947     & 290.874      & 0.745
    \\
    Ours          & \textbf{324.790}     & \textbf{410.314}      & \textbf{0.742} 
                  & \textbf{869.192}     & \textbf{1004.493}    & \textbf{0.823} 
                  & \textbf{572.512}     & \textbf{984.042}     & \textbf{0.765} 
                  & \textbf{135.495}     & \textbf{384.592}     & \textbf{0.762}
    \\
    \bottomrule
    \end{tabular}}
\end{table*}

\subsection{Modality Ablation Study}
~\

\begin{figure}[!t]
        \centering
        \subfigure[Beijing]{
                \includegraphics[width=0.49\linewidth]{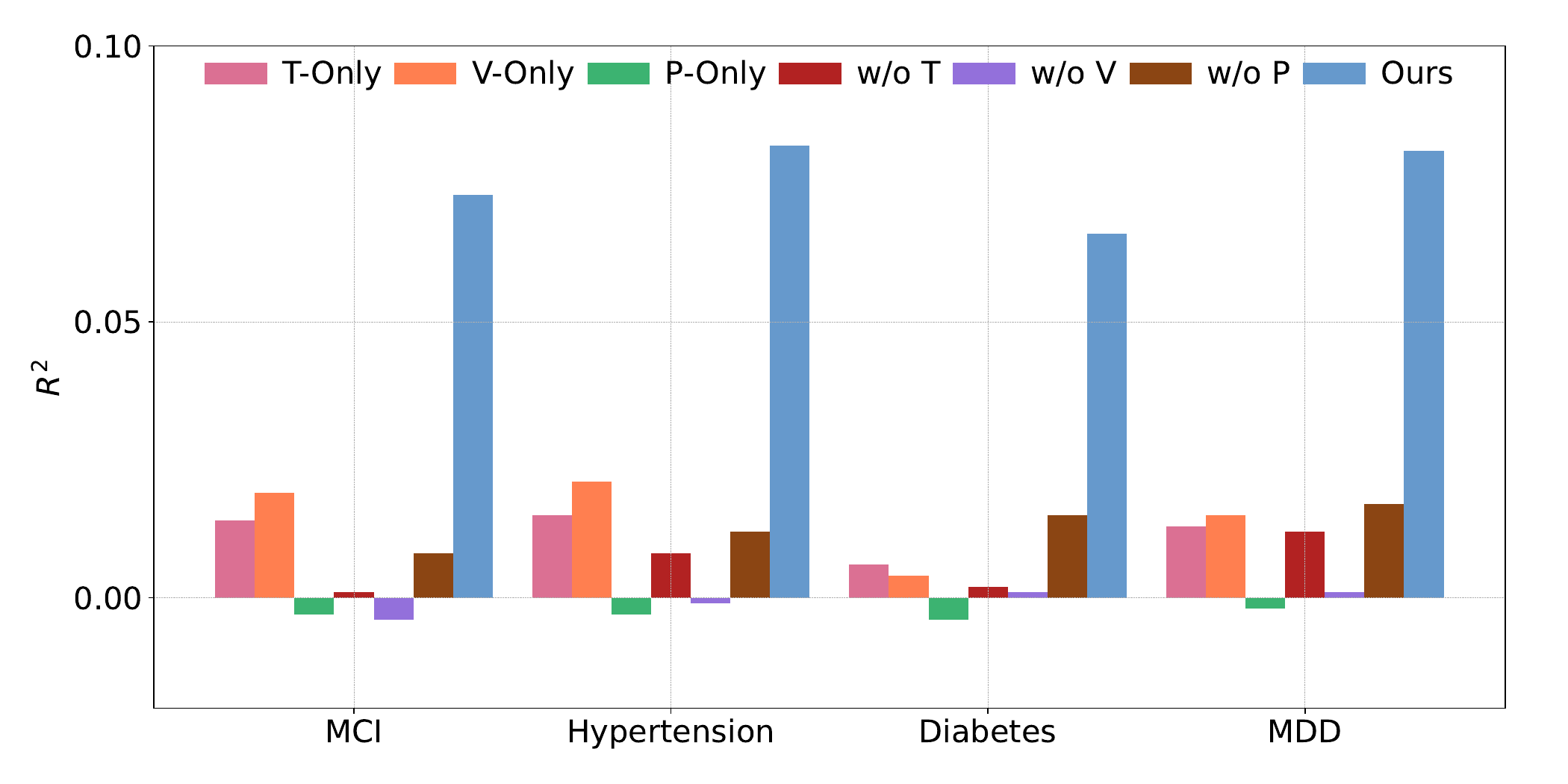}}
        \subfigure[Shanghai]{
                \includegraphics[width=0.49\linewidth]{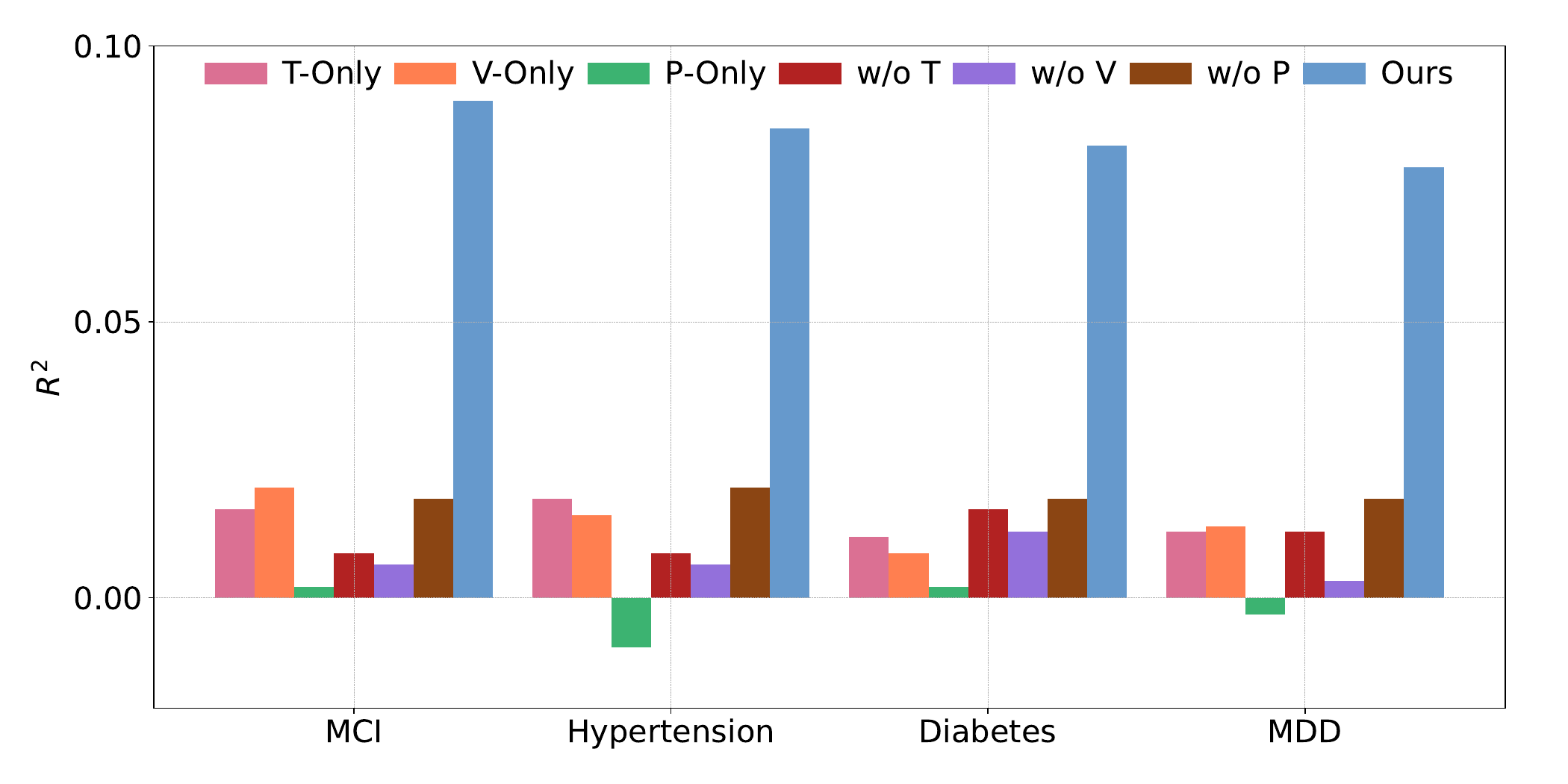}}
        %\vspace{-10px}
        \caption{Ablation study results for $R^2$ in community elder health prediction.}
        \label{fig:Ablation study}
                %\vspace{-10px}
\end{figure}

In this subsection, we conduct a modality ablation study to evaluate the contributions of individual modalities and their combinations within our proposed multi-modal model, CureGraph. Tables~\ref{table:ablation_bj} and~\ref{table:ablation_sh} present CureGraph's performance under various multi-modal configurations for datasets from Beijing and Shanghai, respectively. Additionally, Fig.~\ref{fig:Ablation study} provides a visual representation of the \( R^2 \) values for community elder health predictions in these two cities.

More specifically, the study examines prediction performance across different settings, including single-modality variants (\textbf{T-Only}, \textbf{V-Only}, \textbf{P-Only}) and combined-modality variants (\textbf{w/o T}, \textbf{w/o V}, \textbf{w/o P}). Here, \textbf{T} represents the textual modality, \textbf{V} denotes the visual modality, and \textbf{P} refers to the POI review modality. The results reveal significant variations in predictive performance across these modalities. Notably, the visual modality (V-Only) demonstrates strong predictive capabilities, particularly for MCI and hypertension in Beijing. Furthermore, combining the textual and visual modalities (w/o P) improves predictions for diabetes and MDD, achieving an average improvement of 112.9\% in \( R^2 \) across datasets from Beijing and Shanghai. Conversely, the POI review modality (P-Only) exhibits limited predictive power and does not significantly enhance outcomes when included.

Despite variability in the performance of individual modalities, combining all three modalities consistently yields the highest predictive accuracy (represented by the blue bar for our method in Fig.~\ref{fig:Ablation study}). This improvement can be attributed to the alignment of POI text embeddings with residential textual and visual modalities during multimodal fusion, resulting in better integration within the vector space. These findings emphasize the critical role of multimodal fusion in enhancing predictive accuracy for elderly health conditions.

Additionally, we observe that straightforward combinations of multimodal embeddings, as employed by baseline models like LINE and Node2Vec, perform worse than single-modality approaches. As shown in Table~\ref{table:results_bj}, these baseline models fail to match the performance of single-modality methods across all four chronic disease prediction tasks. This highlights the inadequacy of simplistic multimodal embedding techniques in capturing the complex interactions required for accurate elderly health predictions.

\begin{table*}[ht]
    \centering
    \caption{Ablation study results for community-level elder health prediction in Beijing.}
    \label{table:ablation_bj}
    \resizebox{\textwidth}{!}{
    \begin{tabular}{llllllllllllll}
    \toprule
    \multirow{2}{*}{Model} & \multicolumn{3}{l}{MCI}       & \multicolumn{3}{l}{Hypertension}    &  \multicolumn{3}{l}{Diabetes}   &   \multicolumn{3}{l}{MDD} \\ 
    \cmidrule[0.8pt](lr){2-4}\cmidrule[0.8pt](lr){5-7}\cmidrule[0.8pt](lr){8-10}\cmidrule[0.8pt](lr){11-13}
        & MAE$\downarrow$       & RMSE$\downarrow$     & ${R}^{2}\uparrow$
        & MAE$\downarrow$       & RMSE$\downarrow$     & ${R}^{2}\uparrow$
        & MAE$\downarrow$       & RMSE$\downarrow$     & ${R}^{2}\uparrow$
        & MAE$\downarrow$       & RMSE$\downarrow$     & ${R}^{2}\uparrow$
    \\ \bottomrule
    T-Only 
                 & 21.200      & 31.990     & 0.014         
                 & 26.491      & 39.398     & 0.015
                 & 50.655      & 77.287     & 0.006
                 & 15.718      & 23.406     & 0.013
    \\
    V-Only   
                 & \underline{20.760}          & \underline{31.904}     & \underline{0.019}
                 & \underline{26.017}         & \underline{39.277}     &\underline{0.021}
                 & 50.720          & 77.342     & 0.004
                 & \underline{15.597}          & 23.377     & 0.015
    \\
    P-Only      & 21.486      & 32.250       & -0.003
                & 26.906      & 39.742       &  -0.003
                &  51.282     &  77.651      & -0.004
                & 15.970      & 23.594        & -0.002
    \\       
    w/o T       &  20.982      & 32.198    & 0.001 
                &  26.255      & 39.521     & 0.008 
                & 50.803       & 77.416     & 0.002
                & 15.675       & 23.424     & 0.012
    \\
    w/o V       
                & 21.418  & 32.271   & -0.004 
                & 26.768   & 39.710   & -0.001  
                & 50.893  & 77.483    & 0.001
                & 15.866   & 23.548  & 0.001
    \\
    w/o P       & 21.232     & 32.076    & 0.008
                & 26.573     & 39.433    &  0.012
                & \underline{50.589}     & \underline{76.947}    & \underline{0.015}
                & 15.745    & \underline{23.364}     & \underline{0.017}
     \\ \bottomrule
    Ours         
         &\textbf{20.082}    & \textbf{31.015}   &\textbf{0.073} 
        &\textbf{25.048}   & \textbf{38.023}   &\textbf{0.082} 
        &\textbf{48.552}   &\textbf{74.920}   &\textbf{0.066} 
        &\textbf{14.926}  & \textbf{22.589}   &\textbf{0.081}
    \\ \bottomrule
    \end{tabular}
    }
\end{table*}

\begin{table*}[!ht]
    \caption{Ablation study results for community-level elder health prediction in Shanghai.}
    \label{table:ablation_sh}
    \resizebox{\textwidth}{!}{   
    \begin{tabular}{
        >{\arraybackslash}l
        >{\arraybackslash}l
        >{\arraybackslash}l
        >{\arraybackslash}l
        >{\arraybackslash}l
        >{\arraybackslash}l
        >{\arraybackslash}l
        >{\arraybackslash}l
        >{\arraybackslash}l
        >{\arraybackslash}l
        >{\arraybackslash}l
        >{\arraybackslash}l
        >{\arraybackslash}l
    }
    \toprule
    \multirow{2}{*}{Model} & \multicolumn{3}{l}{MCI}      & \multicolumn{3}{l}{Hypertension}    &  \multicolumn{3}{l}{Diabetes}   &   \multicolumn{3}{l}{MDD}  \\ 
    \cmidrule[0.8pt](lr){2-4}\cmidrule[0.8pt](lr){5-7}\cmidrule[0.8pt](lr){8-10}\cmidrule[0.8pt](lr){11-13}
        & MAE$\downarrow$       & RMSE$\downarrow$     & ${R}^{2}\uparrow$
        & MAE$\downarrow$       & RMSE$\downarrow$     & ${R}^{2}\uparrow$
        & MAE$\downarrow$       & RMSE$\downarrow$     & ${R}^{2}\uparrow$
        & MAE$\downarrow$       & RMSE$\downarrow$     & ${R}^{2}\uparrow$
    \\ \midrule
    T-Only 
                 & 41.786      & 53.229     & 0.016         
                 & 149.803      & 175.873     & 0.018
                 & 64.052      & 72.893     & 0.011
                 & 13.154      & 17.774     & 0.012
    \\
    V-Only   
                 & \underline{41.464}          & \underline{53.067}    & \underline{0.022}
                 & 150.182         & 176.348     & 0.015
                 & 64.219          & 73.006     & 0.008
                 & 13.143          & 17.764     & 0.013
    \\
    P-Only      & 41.796      & 53.904       & 0.002
                & 151.209      & 180.405       &  -0.009
                & 64.486     & 73.041      & 0.002
                & 13.293      & 17.835        & -0.003
    \\       
    w/o T       &  41.871      & 53.580    & 0.008 
                & 151.065      & 177.456     & 0.008 
                & 63.773       & 72.706     & 0.016
                & 13.154       & 17.774     & 0.012
    \\
    w/o V       & 41.846  & 53.688   & 0.006 
                & 151.090   & 177.976   & 0.006  
                & 63.996  & 72.856    & 0.012
                & 13.245   & 17.831  & 0.003
    \\
    w/o P       & 41.625     & 53.203    & 0.018
                & \underline{149.550}     & \underline{175.556}    &  \underline{0.020}
                & \underline{63.147}     & \underline{72.362}    & \underline{0.018}
                & \underline{13.088}   & \underline{17.718}     & \underline{0.018}
     \\ \midrule
    Ours         
         &\textbf{37.568}    & \textbf{48.405}   &\textbf{0.090} 
        &\textbf{138.558}   & \textbf{165.147}   &\textbf{0.085} 
        &\textbf{60.384}   &\textbf{69.087}   &\textbf{0.082} 
        &\textbf{10.567}  & \textbf{16.267}   &\textbf{0.078}
    \\ \bottomrule
    \end{tabular}
}
\end{table*}

\begin{figure}[!t]
        \centering
        \subfigure[Visual]{
                \includegraphics[width=0.32\linewidth]{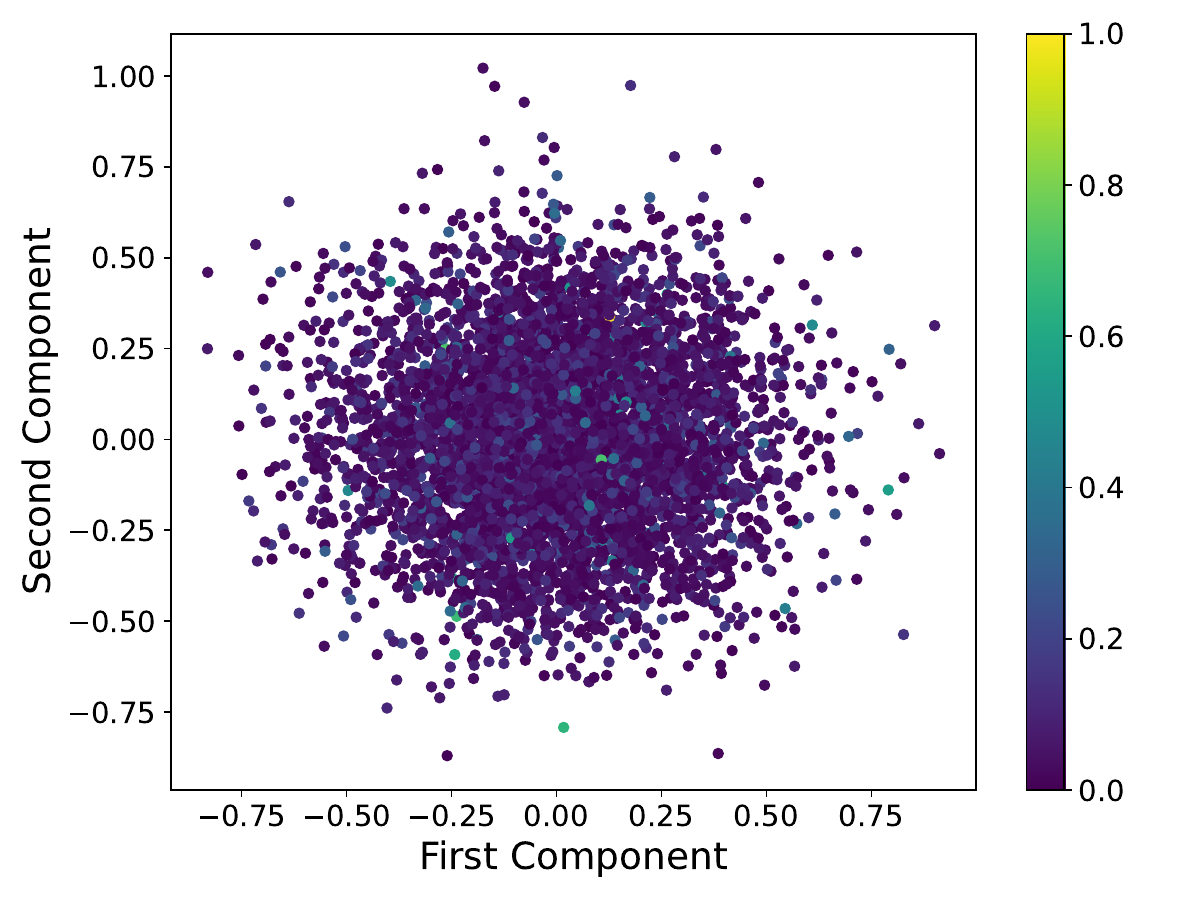}}
        \subfigure[Text]{
                \includegraphics[width=0.32\linewidth]{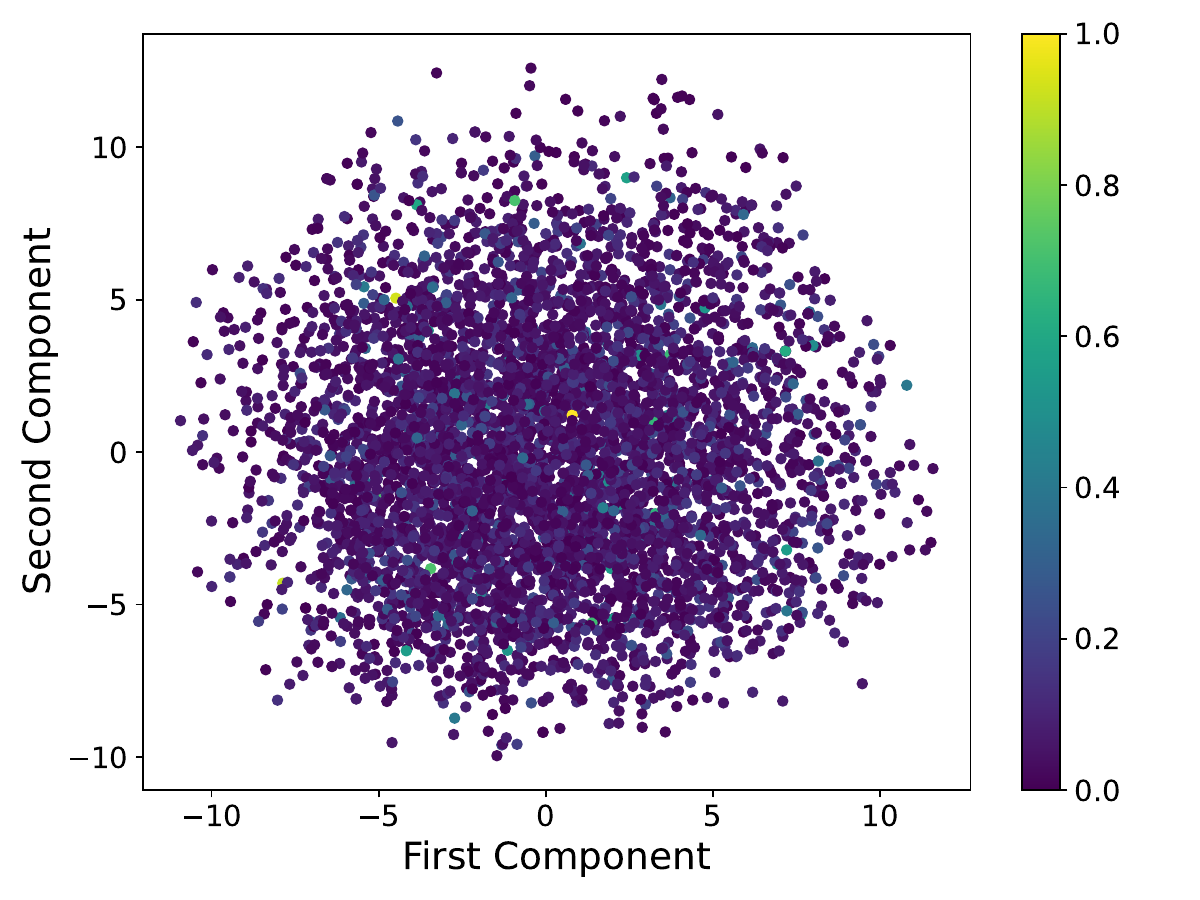}}
        \subfigure[POI review]{
                \includegraphics[width=0.32\linewidth]{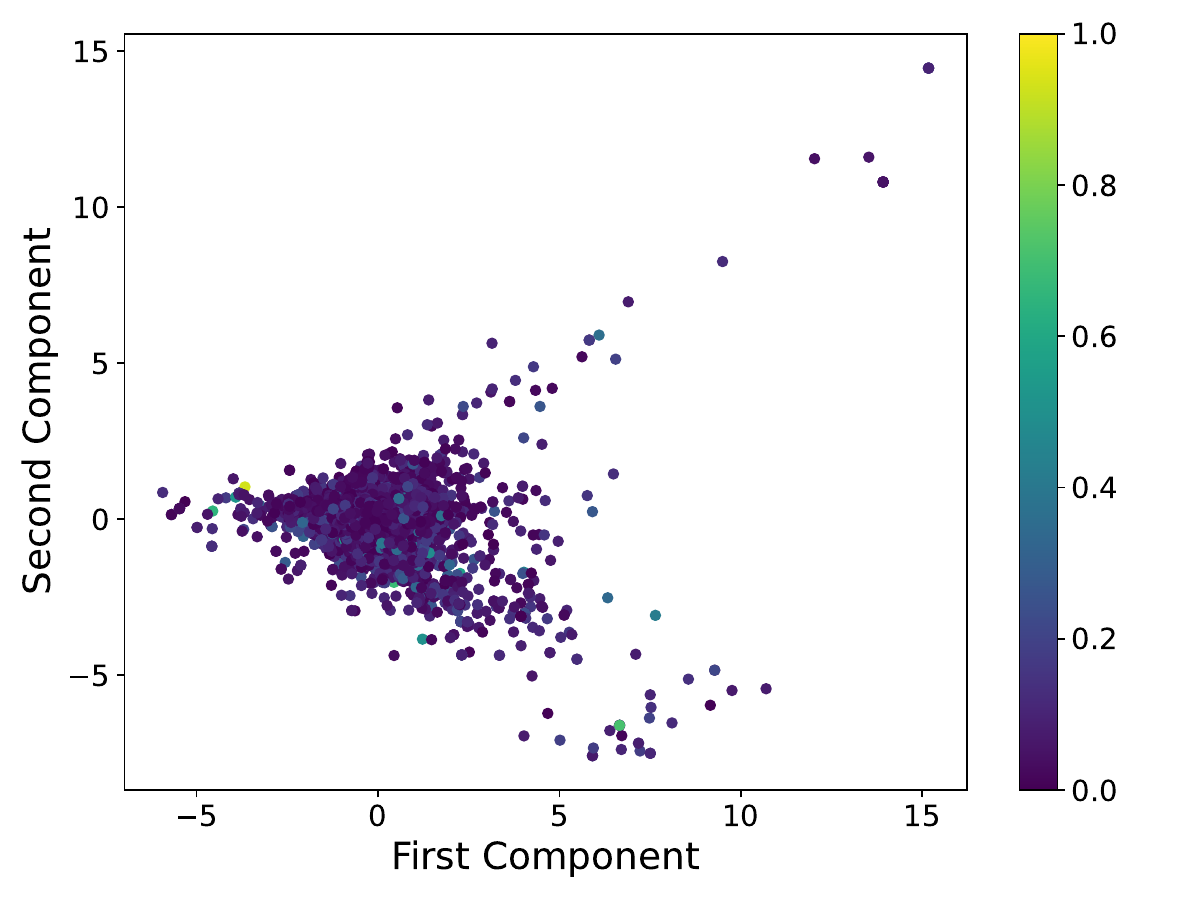}} \\
        \subfigure[Visual]{
                \includegraphics[width=0.32\linewidth]{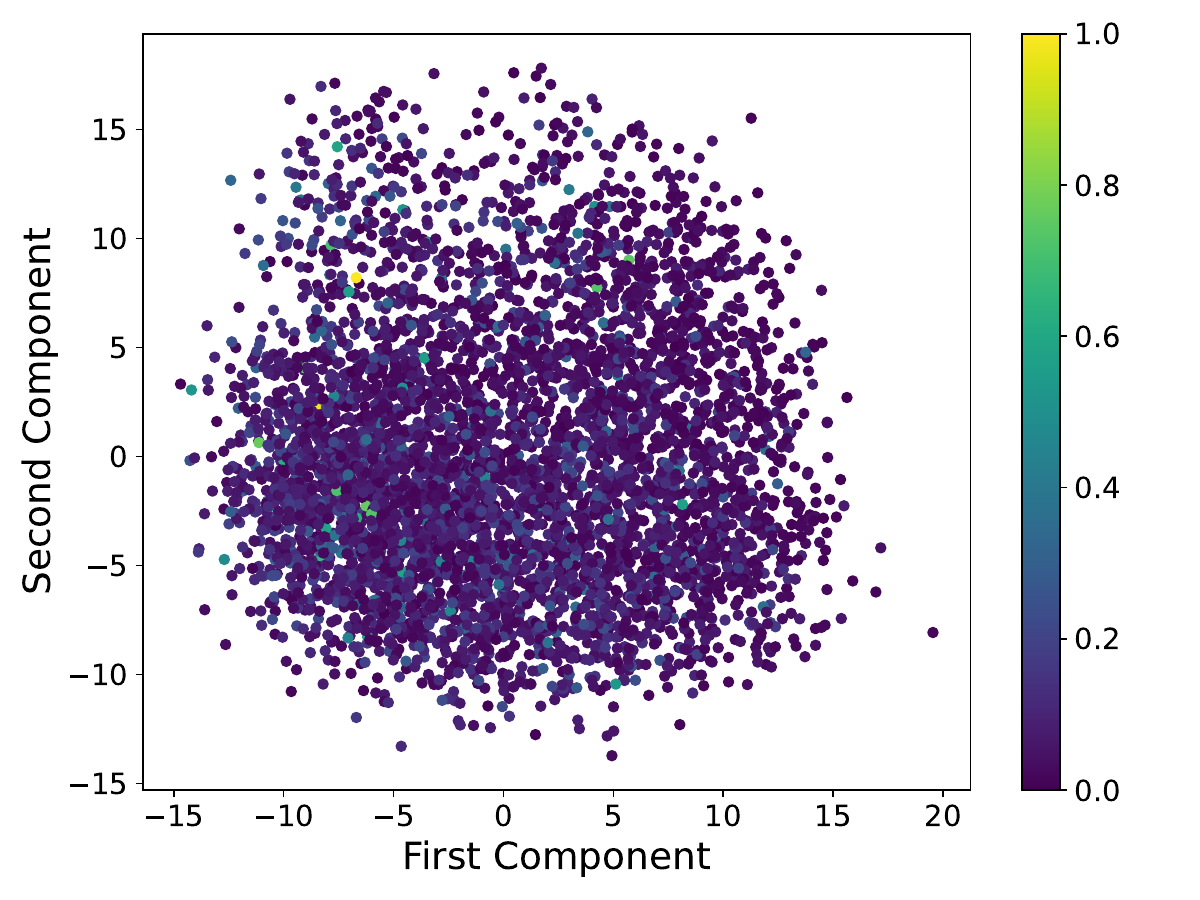}}
        \subfigure[Text]{
                \includegraphics[width=0.32\linewidth]{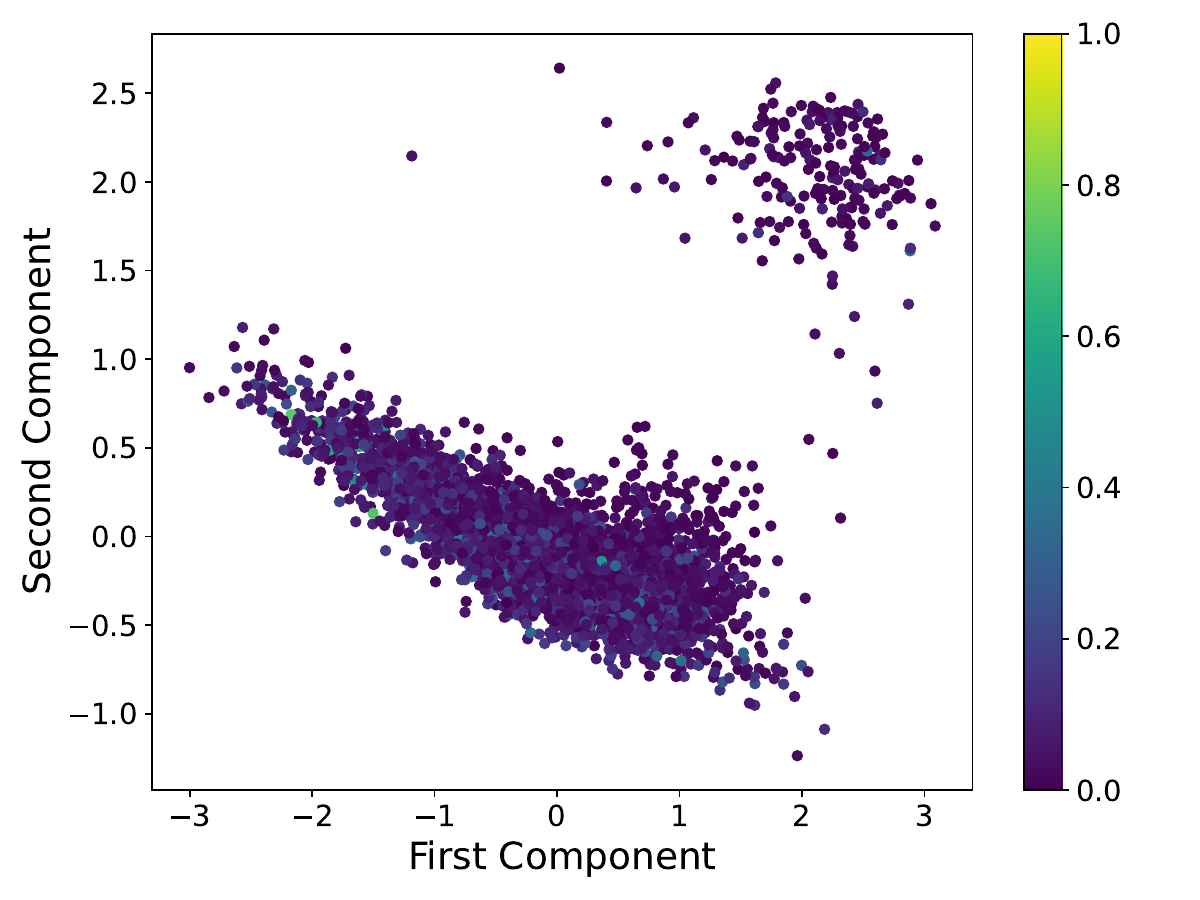}}
        \subfigure[POI review]{
                \includegraphics[width=0.32\linewidth]{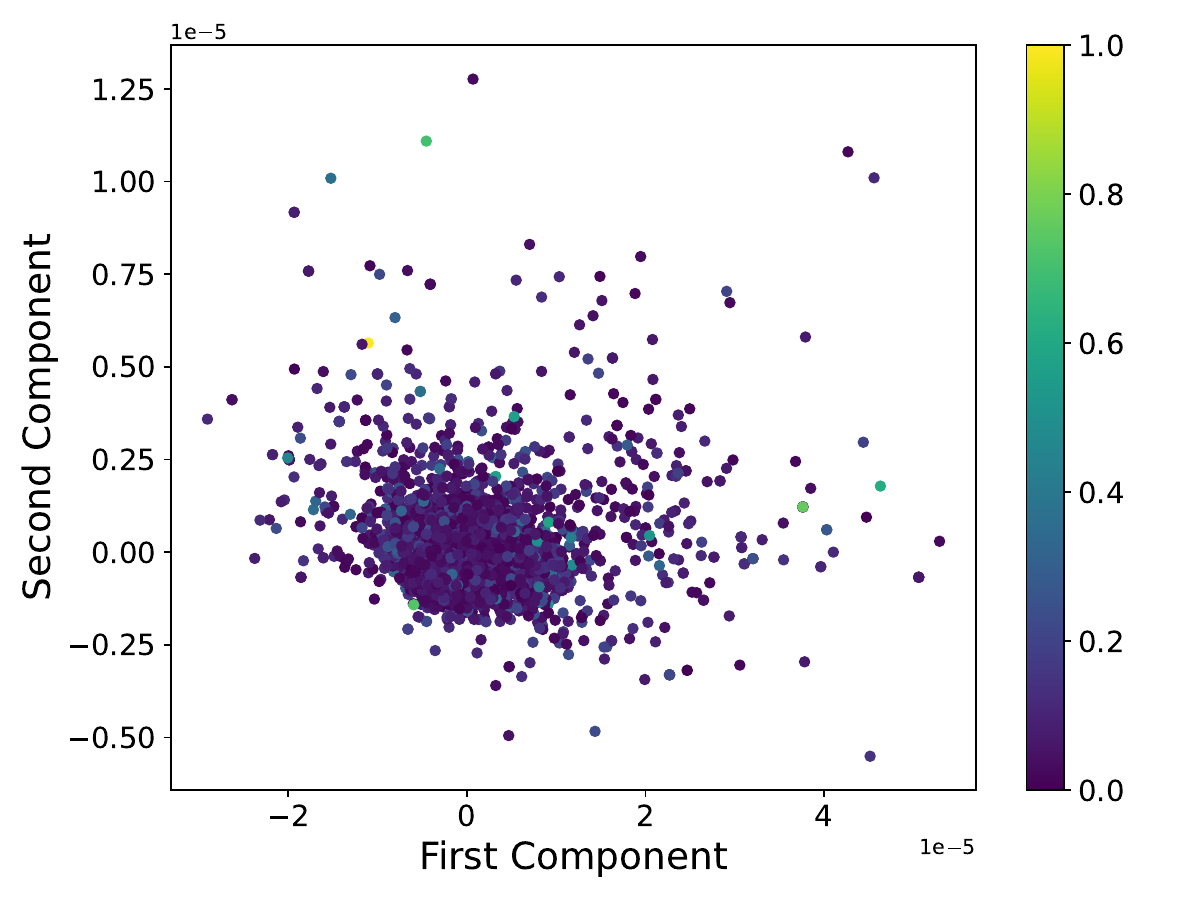}}
        \caption{PCA-based visualization of urban multi-modal embeddings.}
        \label{fig:Visualization}

\end{figure}

\subsection{Analysis of Multi-modal Encoder}
~\

To illustrate the effects of the contrastive learning-based modality encoder and uncover semantic meanings, we visualize the multi-modal embeddings of urban living circles using both pre-trained model methods and our contrastive learning encoder methods. Fig.~\ref{fig:Visualization}(a) and Fig.~\ref{fig:Visualization}(c) employ pre-trained models (DenseNet~\cite{huang2017densely} for images and BERT for texts, respectively), while Fig.~\ref{fig:Visualization}(b) utilizes a self-supervised contrastive learning approach for text encoding. Principal Component Analysis (PCA) is applied to project the features into a two-dimensional feature space, as shown in Fig.~\ref{fig:Visualization}.

The color of each dot corresponds to the population values of geriatric diseases within urban communities. The comparison reveals that the contrastive learning approach for encoding visual and POI text modalities increases the separation in the feature space of living circles, effectively amplifying differences among modalities. Regarding the text modality of residential areas in Fig.~\ref{fig:Visualization}(e), we observe that the first principal component is associated with the number of bus lines, where living circles with multiple bus lines exhibit lower values along this dimension. Furthermore, the text embeddings form two distinct clusters: one primarily associated with residential areas and the other with higher values on both the first and second components, suggesting a strong relationship with transportation and surrounding facilities. Notably, the latter cluster lacks textual descriptions of internal environments, indicating that our modality encoder successfully reduces the distance between semantically similar modalities while increasing the distance between dissimilar ones, particularly when incorporating residential area image semantics through cross-modal contrast learning.

\begin{figure}[!t]
        \centering
        \subfigure[Similarity of reviews based on POI Textual Modality Encoder]{
                \includegraphics[width=0.42\linewidth]{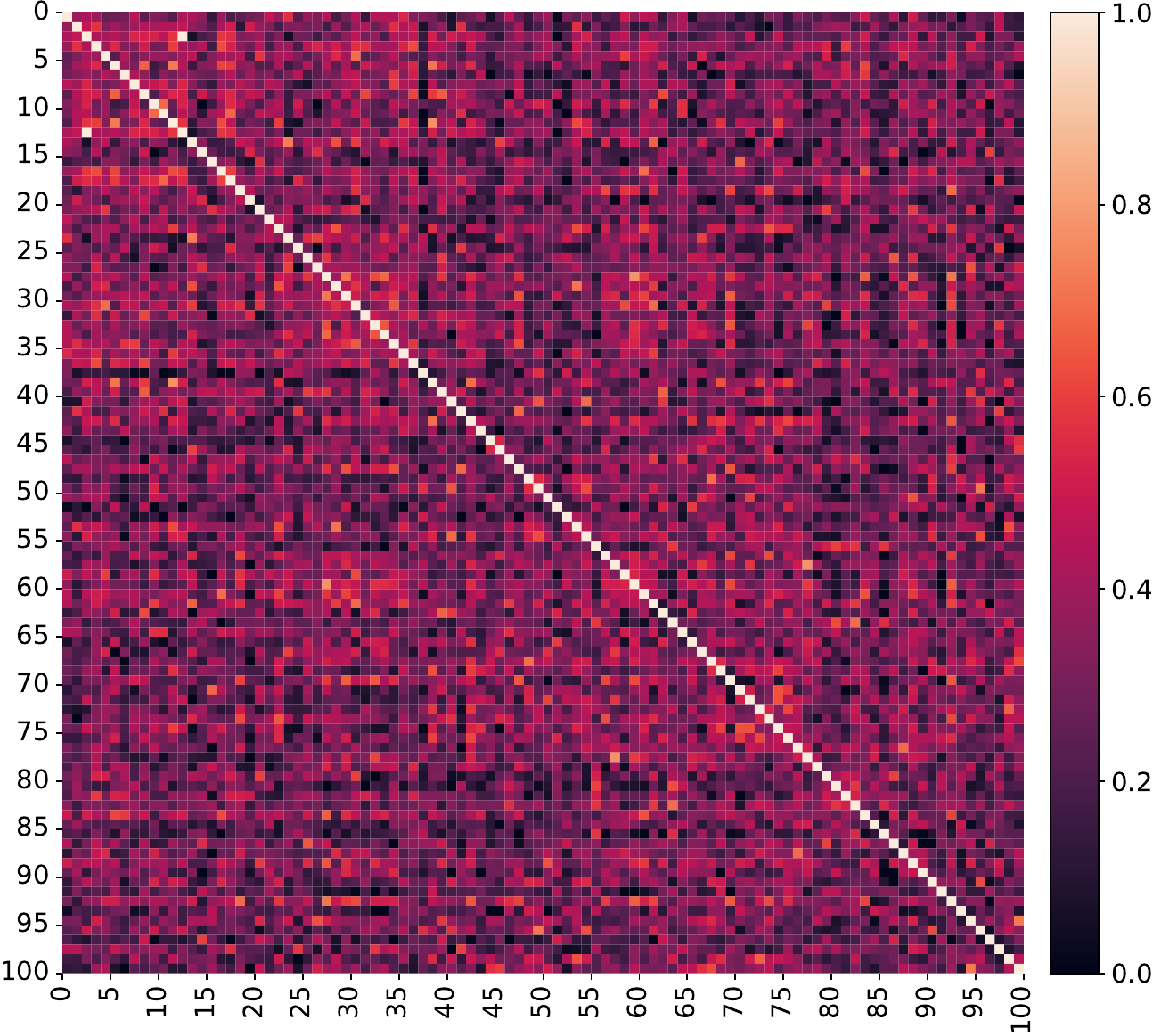}}
        \subfigure[Similarity of reviews based on BERT model]{
                \includegraphics[width=0.42\linewidth]{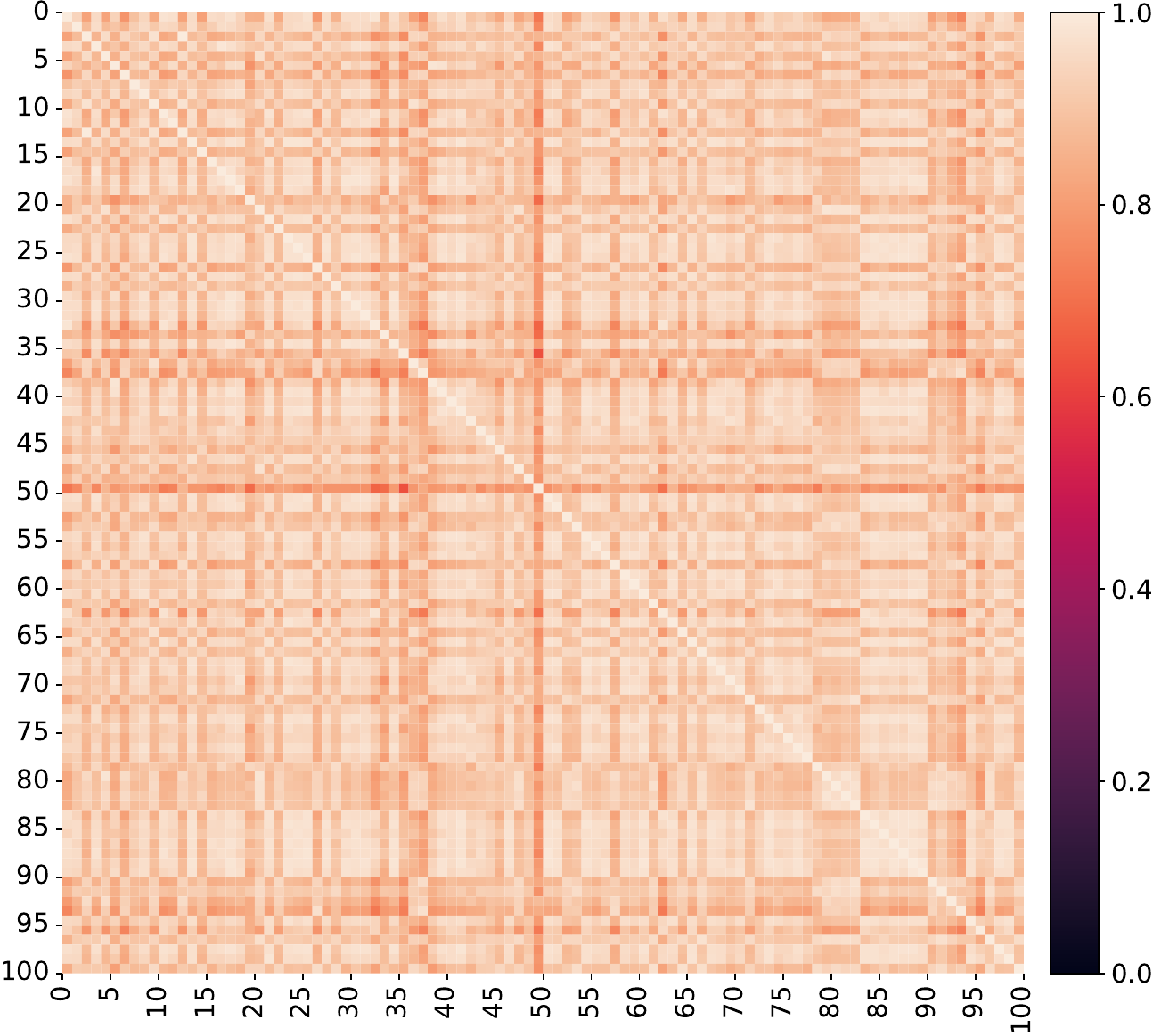}}
        \caption{Heatmaps of POI review representation similarities.}
        \label{fig:Heatmap}
\end{figure}

Furthermore, we compute the pairwise cosine similarities of text representation vectors from POI reviews and present the results as heatmaps in Fig.~\ref{fig:Heatmap}. Specifically, Fig.~\ref{fig:Heatmap}(a) illustrates the results obtained using our proposed POI Textual Modality Encoder, while Fig.~\ref{fig:Heatmap}(b) shows those derived from BERT-based representations of the same texts. For this analysis, we randomly select 100 POI review samples, each comprising 20 reviews sorted by rating (from 1 to 5) in ascending order. Fig.~\ref{fig:Heatmap}(b) demonstrates that the pre-trained language model BERT produces uniformly high textual similarity, with nearly all sentence pairs achieving scores between 0.7 and 1.0. In contrast, Fig.~\ref{fig:Heatmap}(a) reveals that the supervised contrastive learning encoder generates more pronounced disparities between sentence pairs, effectively contracting the similarity of positive samples while expanding the divergence for negative samples. Notably, reviews with lower ratings near the diagonal exhibit distinctly high similarity, underscoring the encoder’s ability to capture subtle distinctions between reviews.

\begin{figure}[!t]
        \centering
        \subfigure[MCI]{
                \includegraphics[width=0.235\linewidth]{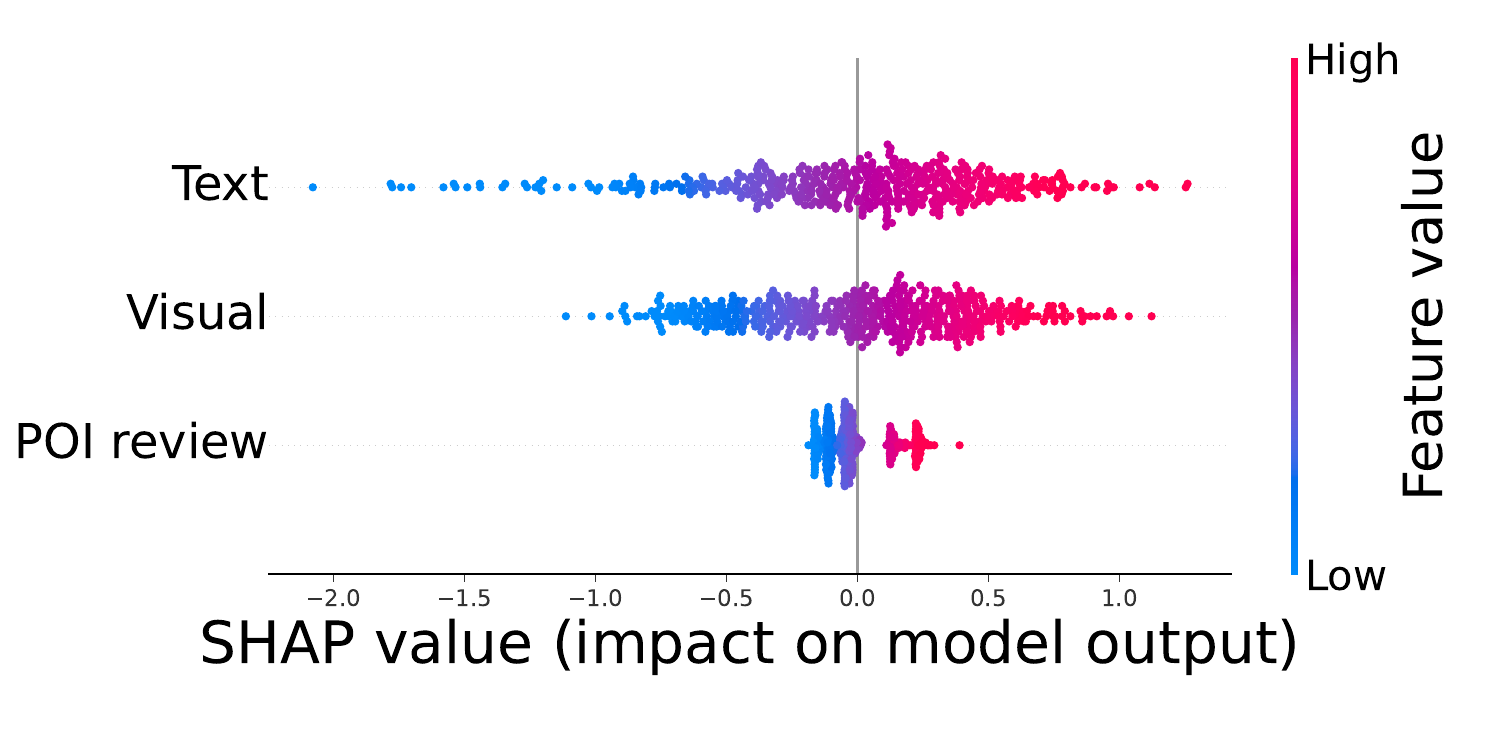}}
        \subfigure[MCI]{
                \includegraphics[width=0.235\linewidth]{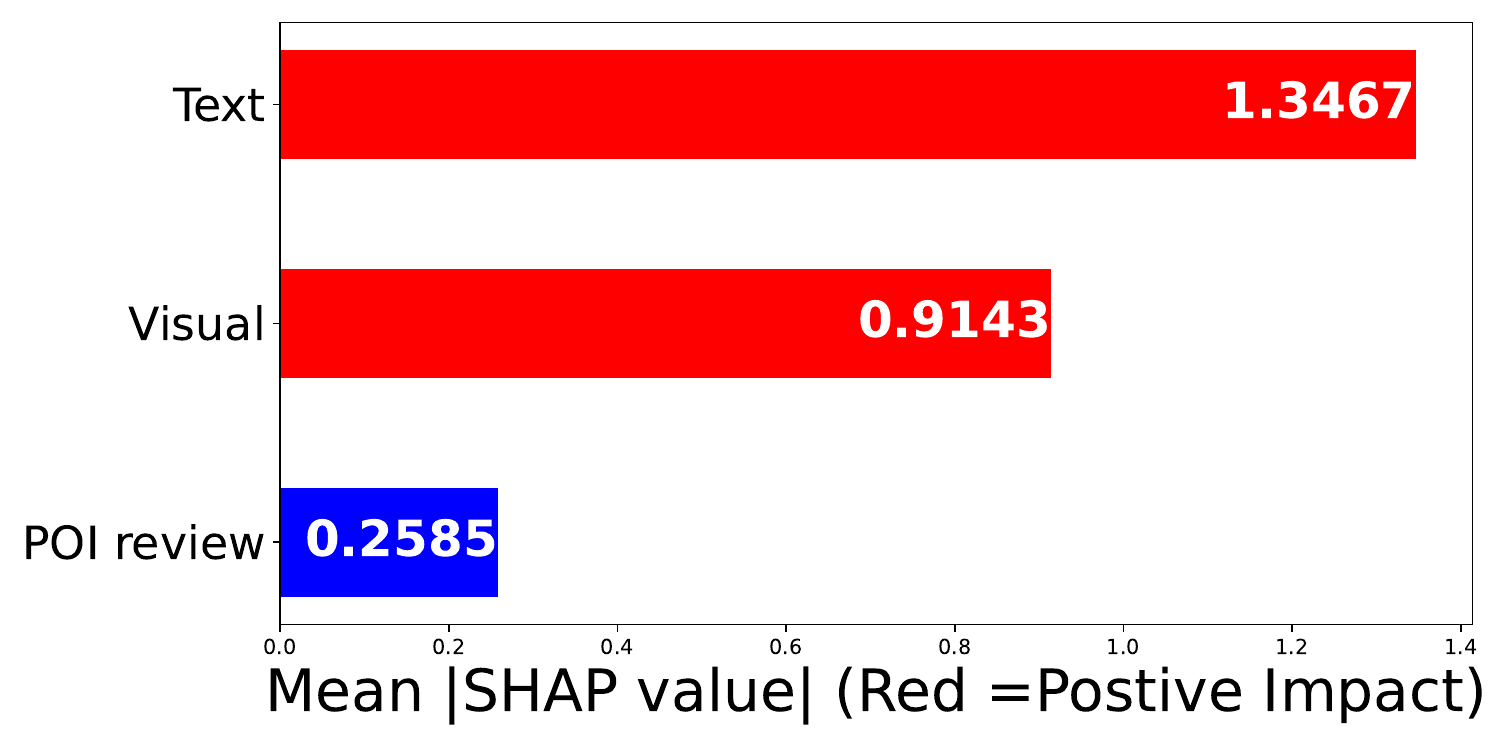}}
        \setcounter{subfigure}{0}
        \subfigure[Hypertension]{
                \includegraphics[width=0.235\linewidth]{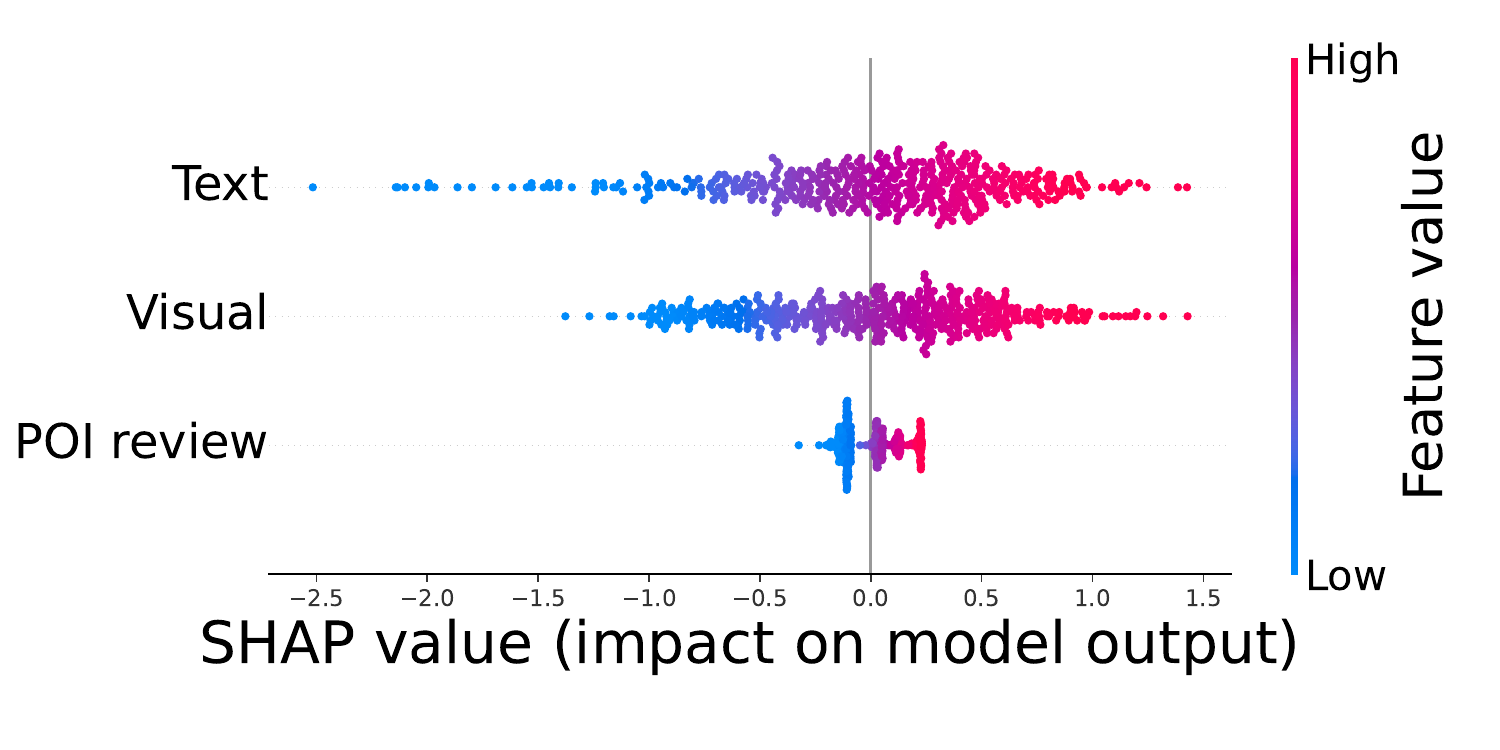}}
        \subfigure[Hypertension]{
                \includegraphics[width=0.235\linewidth]{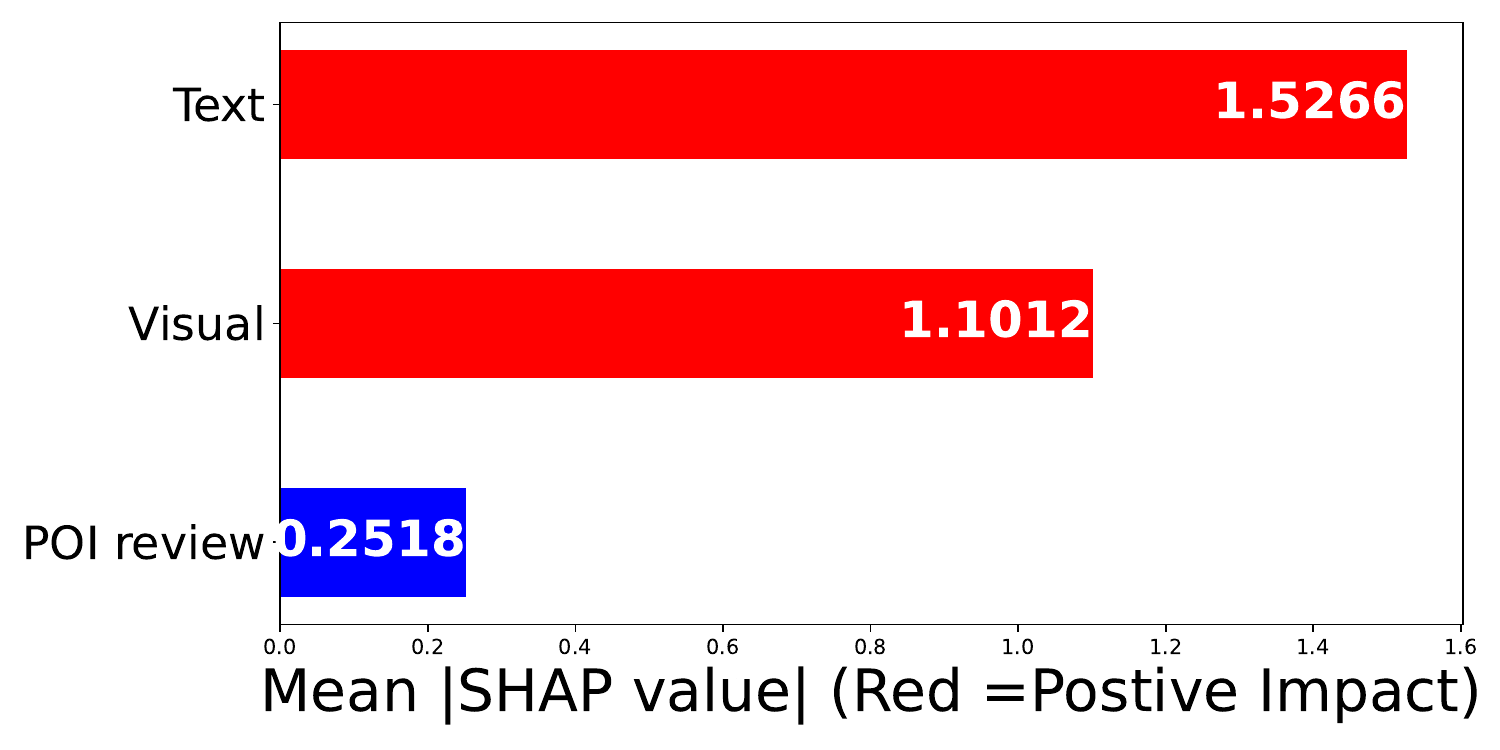}} \\
        \setcounter{subfigure}{0}
        \subfigure[Diabetes]{
                \includegraphics[width=0.24\linewidth]{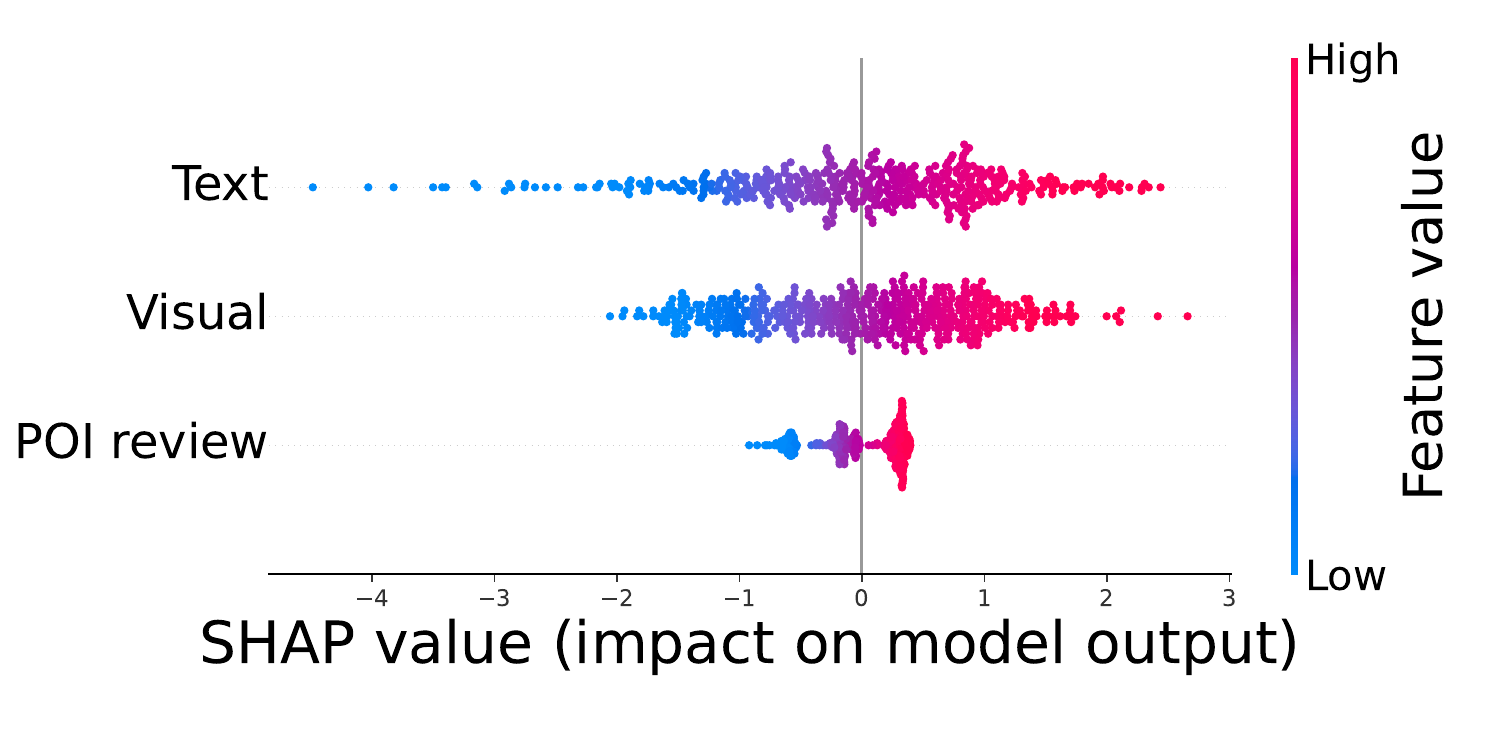}}
        \subfigure[Diabetes]{
                \includegraphics[width=0.24\linewidth]{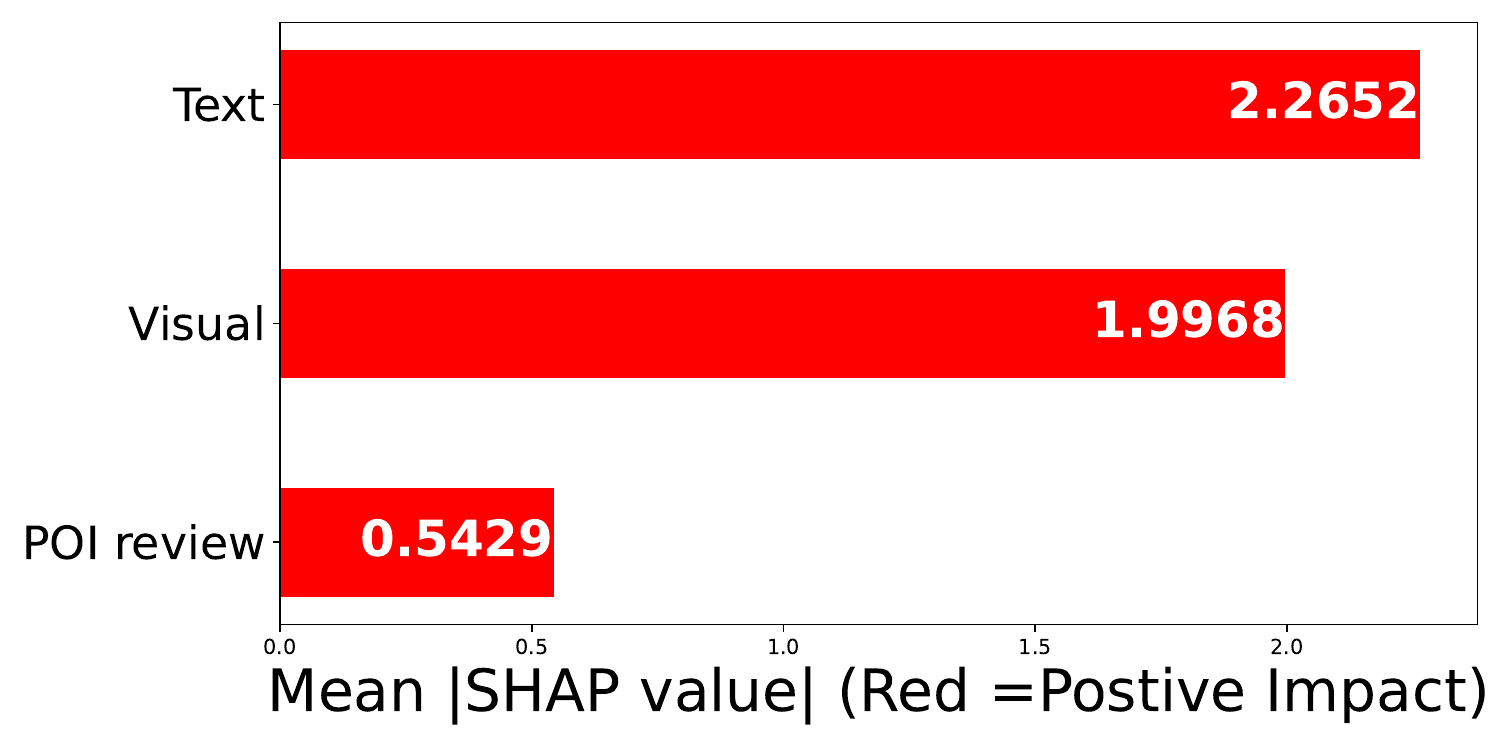}}
        \setcounter{subfigure}{0}
        \subfigure[MDD]{
                \includegraphics[width=0.235\linewidth]{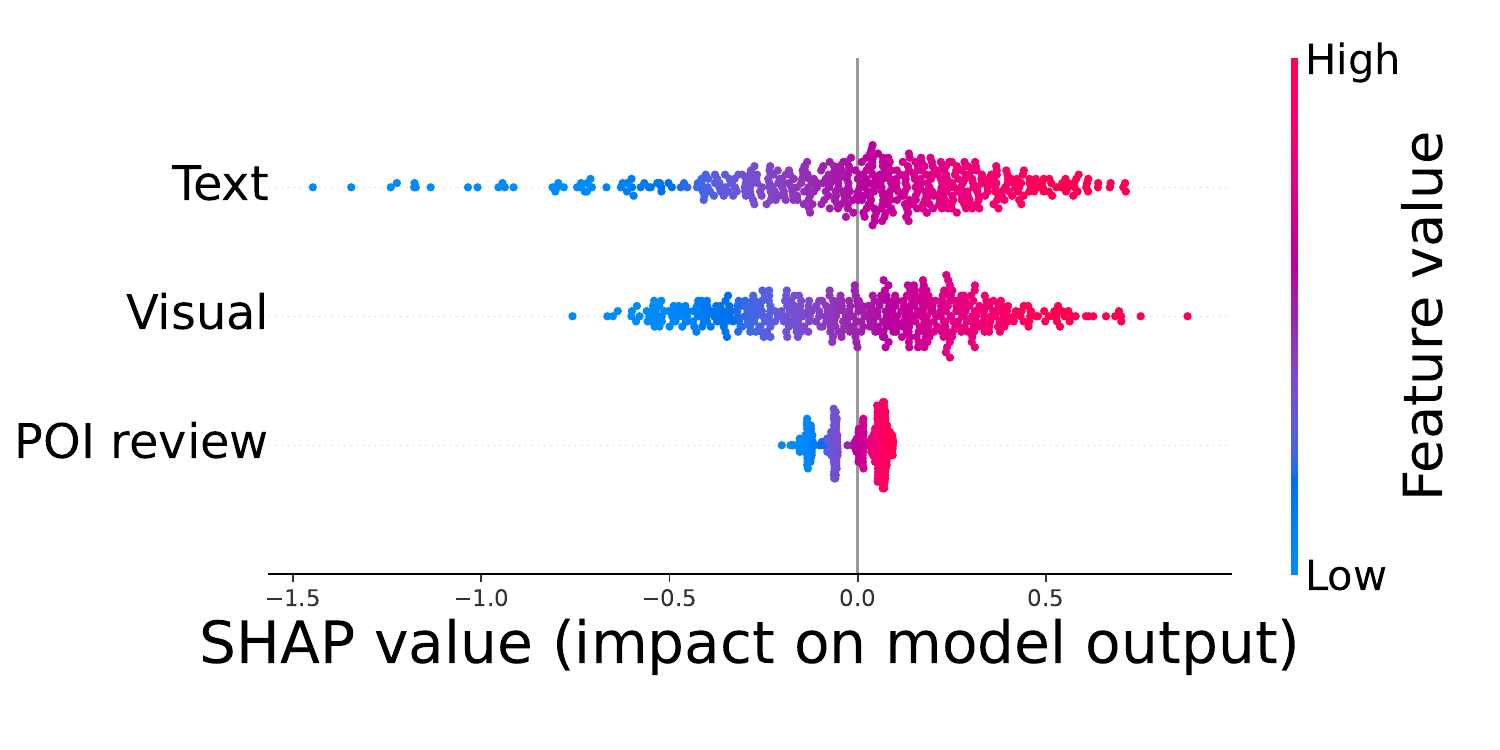}}
        \subfigure[MDD]{
                \includegraphics[width=0.235\linewidth]{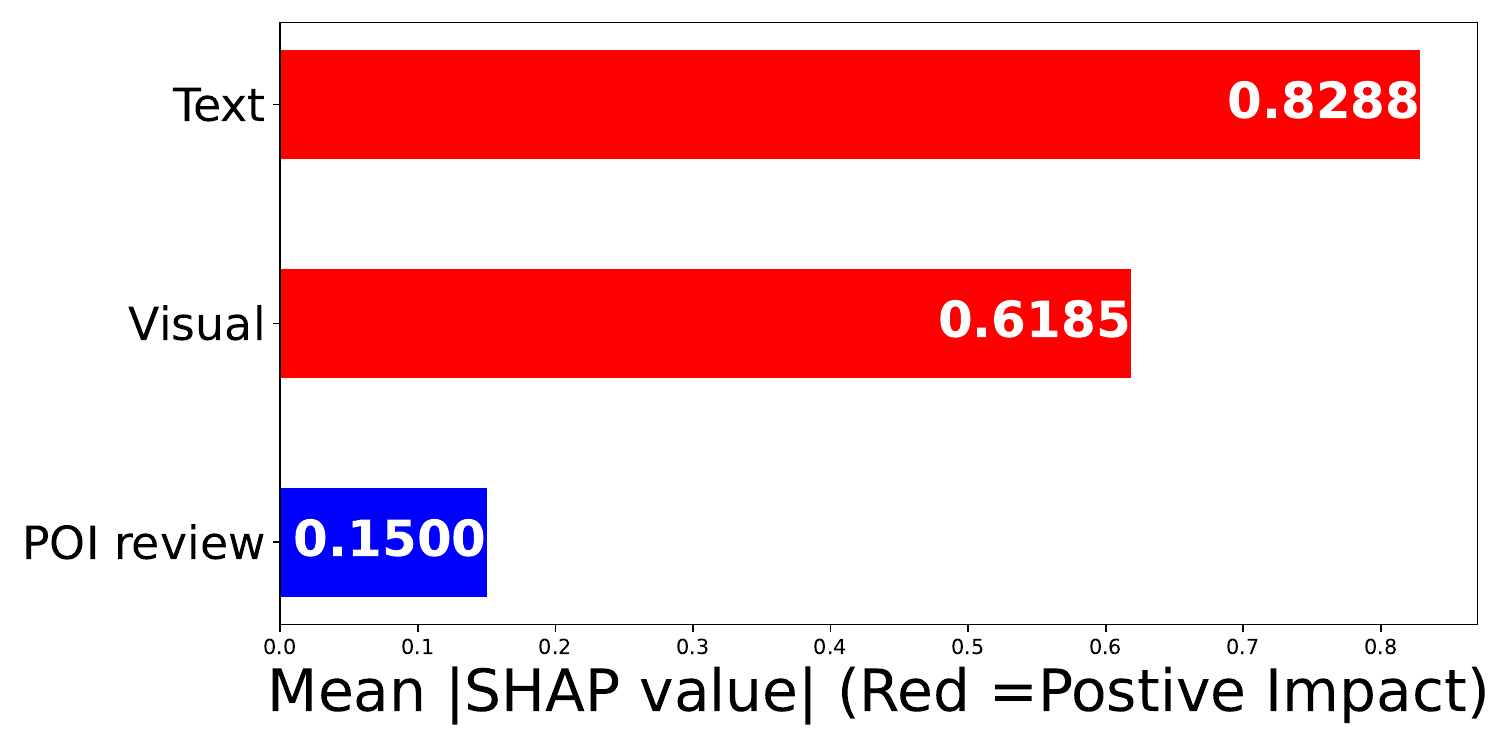}}
        \caption {SHAP value plots for prediction analysis. (a) A summary of SHAP values for modal embeddings. (b) Average absolute SHAP values, with color indicating positive (red) or negative (blue) impacts on the prediction outcomes.}
        \label{fig:shap}
\end{figure}

\subsection{Analysis of Modality Impact}
~\

To better understand the influence of multimodal embeddings on the prediction outcomes for elderly diseases, we employ Shapley Additive Explanations (SHAP) values~\cite{lundberg2020local} to evaluate the importance of various features. SHAP values quantify the contribution of individual features to specific predictions, providing deeper insights into how different features derived from multiple modalities affect CureGraph's outputs. To ensure computational efficiency, we randomly select embedding data from 500 communities in Beijing to assess their roles in the regression model's predictions.

In Fig.~\ref{fig:shap}, panel (a) illustrates the SHAP values for different modalities in predicting each elderly disease. Modal features are represented as horizontal lines, with colors indicating the magnitude of feature values: red points denote higher feature values, while blue points represent lower values. Panel (b) shows the average absolute SHAP value for each feature, along with the correlation between feature values and their SHAP values. Features depicted by red bars have positive correlation coefficients, meaning that higher feature values are associated with an increased predicted number of elderly individuals with the disease. Conversely, features shown with blue bars exhibit negative correlations, where higher feature values correspond to a decreased predicted number of elderly individuals with the disease.

Among the four diseases analyzed, the text modality learned by CureGraph has the most significant impact on prediction outcomes, followed by the visual modality, with POI review data having the least influence. Interestingly, compared to the other two modalities, higher POI review feature values are generally positively correlated with the number of elderly individuals suffering from MCI, hypertension, and MDD in the community, but negatively correlated with the number of elderly individuals suffering from diabetes.

\subsection{Analysis of Health Disparities}
~\
\subsubsection{Health Differences at the Community Level}

\begin{figure}[htb]
	\centering
	\includegraphics[width=0.8\textwidth]{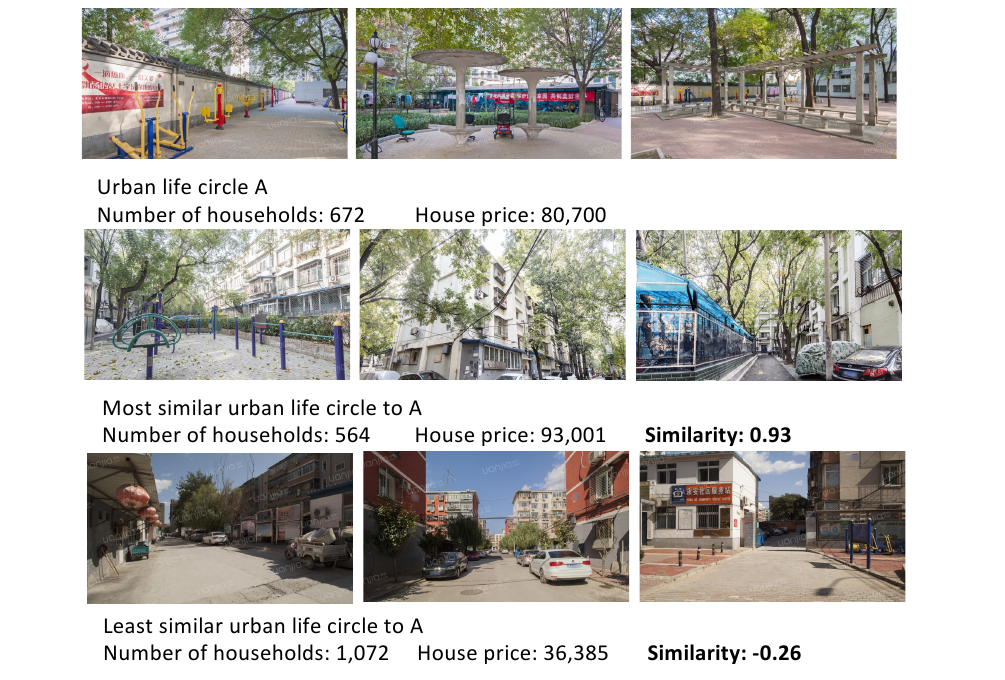}
    \caption{Community health differences in Beijing based on living circle embeddings.}
	%\vspace{-2ex}
	\label{fig:Community_Health}
	\vspace{-2ex}
\end{figure}

For a given living circle within a city, we compute the cosine similarity between its embedding vector and those of other neighborhoods in the city. Based on these similarity metrics, we rank the neighborhoods. In Fig.~\ref{fig:Community_Health}, we present two examples of neighborhood comparisons with cosine similarity scores of 0.93 and -0.26, respectively.

As shown in Fig.~\ref{fig:Community_Health}, urban living circle A, presented in the first row, and its most similar neighborhood (second row) exhibit striking resemblances in physical appearance and living environment. Both areas are characterized by convenient transportation, low floor area ratios, abundant green spaces, and high housing prices. Additionally, they share similarities in the number of residential households and the distribution of the elderly population. In contrast, the community least similar to urban living circle A (third row) displays distinct features, such as lower housing values, limited green spaces, and significant challenges related to aging. This community faces a substantial elderly population combined with a low coverage rate of relevant service facilities.

\subsubsection{Health Differences at the Street Level}
~\

\begin{figure}[htb]
        \centering
        \subfigure[K-means clustering of street embeddings in Beijing(K=3)]{
                \includegraphics[width=0.45\linewidth]{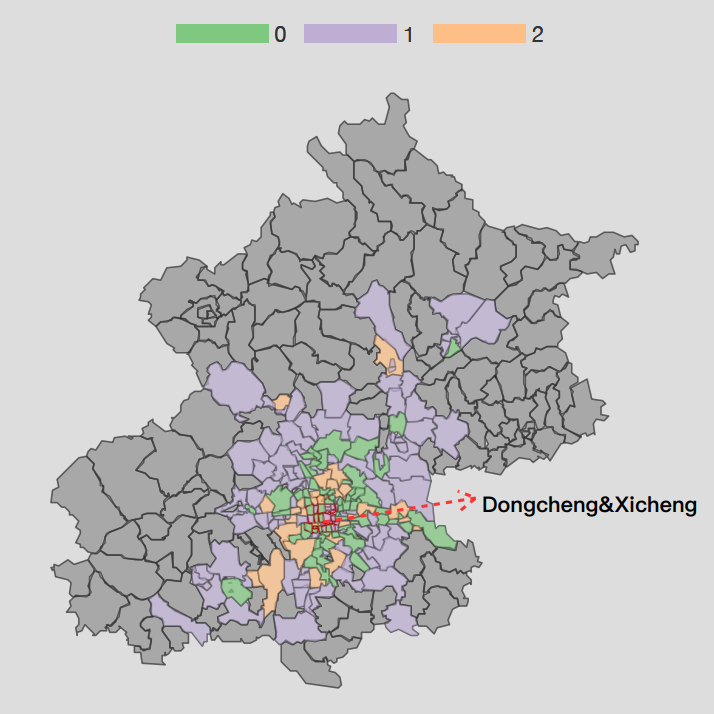}}
        \subfigure[Distribution of the population aged 60 and above in Beijing]{
                \includegraphics[width=0.45\linewidth]{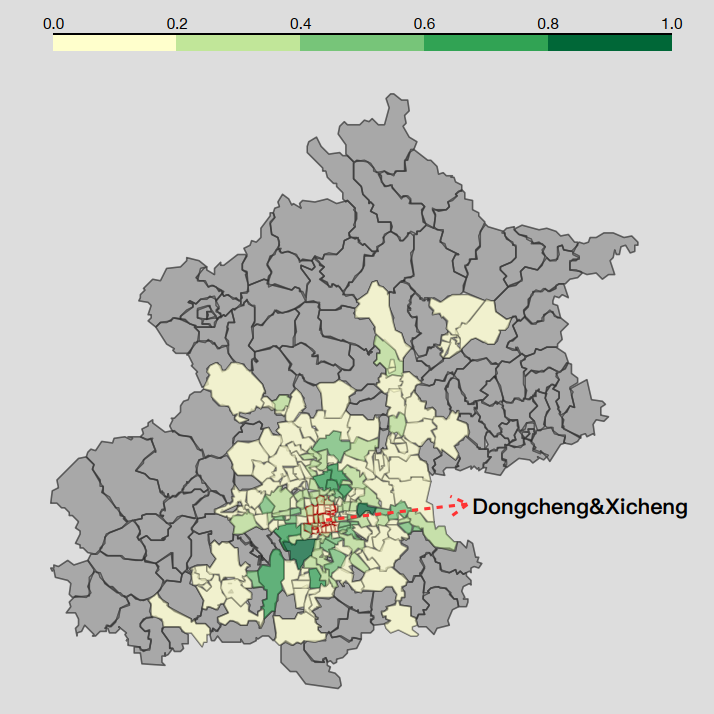}}
        \caption{Cluster and choropleth maps of street embeddings and elderly population in Beijing.}
        \label{fig:street_map}
\end{figure}

To evaluate health disparities at the street level, we apply K-means clustering on embeddings generated by our model for Beijing. Fig.~\ref{fig:street_map}(a) displays the clustering results with \(k=3\), determined by the elbow method. Fig.~\ref{fig:street_map}(b) depicts the distribution of the population aged 60 and above in Beijing. As shown in the figure, a notable cluster 0 (green) emerges around streets with high facility coverage outside the city's second ring road and in certain suburban areas. These regions are characterized by high population mobility, frequent activities, and dense urban development. Conversely, cluster 1 (purple) encompasses sparsely populated areas, indicating low concentrations of elderly individuals. Within the second ring road, the Dongcheng and Xicheng districts exhibit a distinct clustering pattern, underscoring a significant aging demographic challenge. These central areas are outlined by a red-bordered boundary. Streets in the orange cluster predominantly represent extensive, underdeveloped township administrative regions, discernible from their delineated boundaries. Fig.~\ref{fig:street_map}(b) further highlights that these regions contain the highest concentrations of elderly individuals.

Our analysis indicates comparable health representations and elderly population distributions between suburban and central urban areas. However, Beijing municipal census data reveals a significant aging challenge in the city center. In districts such as Dongcheng and Xicheng, the proportion of the population aged 60 and above reaches approximately 26\%. These central communities urgently require elderly-friendly facilities and supportive welfare policies to address the needs of their aging residents.

\subsubsection{Socioeconomic Bias}
~\

To investigate potential socioeconomic biases, we conduct a correlation analysis examining the relationship between predicted elderly disease prevalence and socioeconomic factors. Using the Pearson correlation coefficient (PCC)~\cite{cohen2009pearson} and p-value, we assess the association between predicted disease prevalence and housing prices. The results indicate a significant negative correlation (\( \text{PCC} = -0.17, p < 0.05\)), suggesting that communities with higher housing prices tend to have fewer predicted elderly individuals with diseases.
This finding may reflect biases in the data collection process. Communities with higher housing prices often feature higher-quality images and more detailed textual descriptions, which can enhance the model’s predictive performance in these areas. Conversely, economically disadvantaged communities may suffer from lower data quality or insufficient details, resulting in relatively less accurate predictions. These disparities highlight the importance of addressing data quality issues to ensure equitable model performance across different socioeconomic contexts.

\section{Conclusion}
~\

In this work, we introduce \textbf{CureGraph}, an innovative unsupervised graph-based multi-modal representation learning framework that leverages photos, textual reviews of residential areas, and surrounding POI data to generate urban neighborhood health embeddings. Our framework constructs a graph to capture cross-modal dependencies and spatial autocorrelation among neighborhoods, enabling the effective fusion of diverse data modalities. This approach allows for fine-grained predictions of community elder health statuses and uncovers urban health disparities across multiple spatial levels. We validate CureGraph through experiments on real-world datasets and extensive ablation studies under various modality settings. The findings offer valuable insights for urban planners and policymakers, fostering healthier urban communities and supporting spatial planning for healthy aging.

Despite its strengths, our approach has certain limitations. First, it relies predominantly on static images and textual data, limiting its ability to capture the dynamic nature of urban environments and their continuous evolution. Second, while our framework accounts for spatial autocorrelation among urban regions, it does not fully incorporate road network data. This is a significant limitation, as road networks play a crucial role in revealing intricate connections and interaction patterns within cities, profoundly influencing mobility, lifestyle behaviors, and community interactions.

To address these limitations, future work will focus on incorporating dynamic data sources to better capture urban vibrancy and change. We also plan to integrate road network data into the framework to gain a deeper understanding of urban spatial dynamics and complexities. To enhance the inclusiveness and precision of CureGraph, we aim to design novel algorithms that can adapt to evolving urban contexts. Furthermore, with a strong emphasis on privacy protection, we will explore anonymized or pseudo-anonymized data collection methods and collaborate with healthcare institutions and community service organizations. Accessing mobility and health data specific to the elderly population will enable the provision of more targeted and effective services with greater precision.
These advancements are expected to significantly enhance insights into urban planning and governance, promoting environments that support healthy aging and fostering improved community welfare.

\section*{Acknowledgement}
This work was supported by the National Natural Science Foundation of China (NSFC, Grant No. 62106274) and the Fundamental Research Funds for the Central Universities, Renmin University of China (Grant No. 22XNKJ24).

\bibliographystyle{plainnat}
\bibliography{references}

\end{document}